\documentclass[sigconf]{acmart}
\AtBeginDocument{%
  }

\usepackage{makecell}
\usepackage{enumitem}
\usepackage{multirow}
\usepackage{subcaption}
\usepackage[capitalize,noabbrev]{cleveref}

\newcommand{\weeny}{\fontsize{4pt}{4.5pt}\selectfont}
\newcommand{\teenyweeny}{\fontsize{3pt}{3.5pt}\selectfont}
\newcommand{\squeezegap}{\vspace{-12pt}}
\copyrightyear{2026}
\acmYear{2026}
\setcopyright{cc}
\setcctype{by-nc-nd}
\acmConference[ICMR '26]{International Conference on Multimedia Retrieval}{June 16--19, 2026}{Amsterdam, Netherlands}
\acmBooktitle{International Conference on Multimedia Retrieval (ICMR '26), June 16--19, 2026, Amsterdam, Netherlands}
\acmDOI{10.1145/3805622.3810794}
\acmISBN{979-8-4007-2617-0/2026/06}

\begin{document}

\title{CompArt: Operationalizing Aesthetic Alignment in Text-to-Image Generation via Principles of Art}


\author{Zhe JIN}
\email{jinzhe@u.nus.edu}
\orcid{0000-0002-9469-7720}
\affiliation{%
  \institution{National University of Singapore}
  \country{Singapore}
}

\author{Tat-Seng CHUA}
\email{dcscts@nus.edu.sg}
\orcid{0000-0001-6097-7807}
\affiliation{%
  \institution{National University of Singapore}
  \country{Singapore}
}

\renewcommand{\shortauthors}{Jin et al.}

% Abstract: 1/4 page

\begin{abstract}
Text-to-Image (T2I) diffusion models have made rapid progress on \emph{semantic alignment} (generating \emph{what} is described in the prompt), yet users still lack reliable control over \emph{aesthetic composition} (how visual elements are put together). Prior work often treats aesthetics as a single, preference-driven notion (e.g., ``high quality'', ``detailed'', ``breathtaking''), which does not map cleanly to compositional intent.

We propose \textit{Aesthetic Alignment}: aligning generated images to \emph{explicit, user-specified compositional constraints}. We operationalize these constraints using the Principles of Art (PoA)—e.g., Balance, Rhythm, and Emphasis—commonly used in art education to describe composition. To support this task, we introduce \textbf{CompArt}, a dataset of 80,032 WikiArt images augmented with captions and PoA analyses produced by a multimodal LLM under structured prompting. We further propose \textbf{ArtDapter}, a lightweight and disentangled adapter that enables steering a pretrained T2I model along 10 PoA dimensions while retaining the base model’s semantic capability. Experiments on CompArt show improved adherence to PoA controls over strong baselines under a dual evaluation protocol.
\end{abstract}




\begin{teaserfigure}
  {\scriptsize
    \def\colw{.19\textwidth}
    \def\imgw{.8\linewidth}
    \def\txtw{.9\linewidth}
    \setlength{\columnsep}{0pt}
    %
    % --- ROW 1 ---
    \begin{minipage}[t]{\colw}
        \centering
        \includegraphics[width=\imgw]{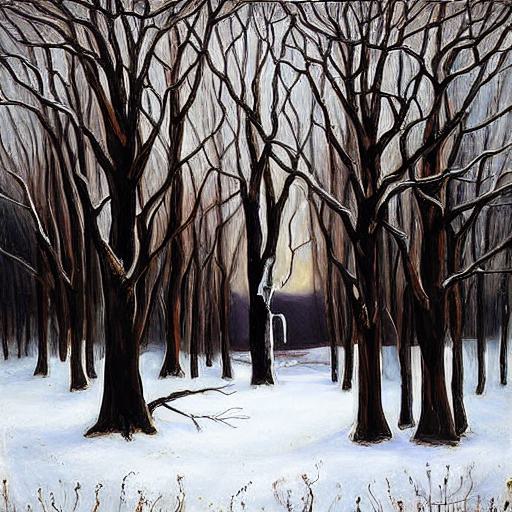}
        \parbox[t]{\txtw}{\raggedright
        \textbf{Context}: A snowy forest scene with bare trees surrounding a dark, frozen pond.\\
        \textbf{Art-Style}: Post-Impressionism.\\
        \textbf{Symmetric balance} is evident in the composition, with the dark pond centrally placed and flanked by similar arrangements of snow-covered trees on both sides.}
    \end{minipage}\hfill
    \begin{minipage}[t]{\colw}
        \centering
        \includegraphics[width=\imgw]{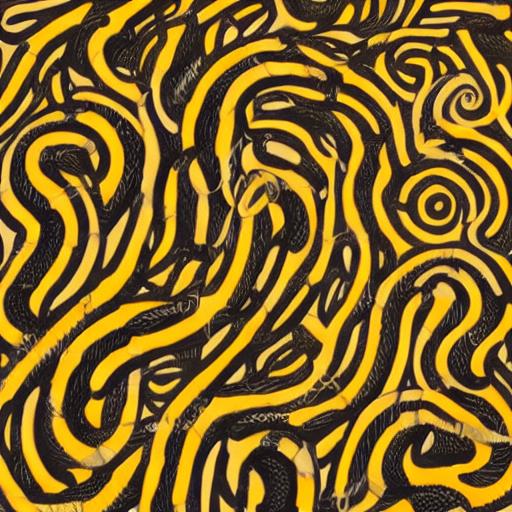}
        \parbox[t]{\txtw}{\raggedright
        \textbf{Context}: A coiled snake with a detailed pattern on its body, against a plain background with handwritten text.\\
        \textbf{Art-Style}: Symbolism.\\
        \textbf{Harmony} is achieved through the consistent color palette of browns, blacks, and yellows, which unifies the snake with the background.}
    \end{minipage}\hfill
    \begin{minipage}[t]{\colw}
        \centering
        \includegraphics[width=\imgw]{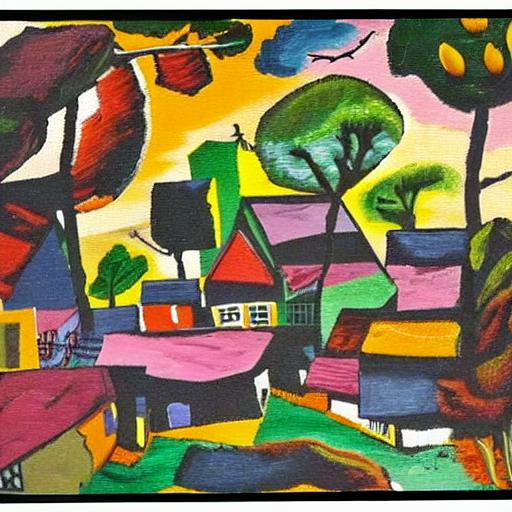}
        \parbox[t]{\txtw}{\raggedright
        \textbf{Context}: A vibrant scene of houses surrounded by trees with a colorful sky in the background.\\
        \textbf{Art-Style}: Expressionism.\\
        \textbf{Variety} is present in the diverse range of colors, shapes, and textures used throughout the composition.}
    \end{minipage}\hfill
    \begin{minipage}[t]{\colw}
        \centering
        \includegraphics[width=\imgw]{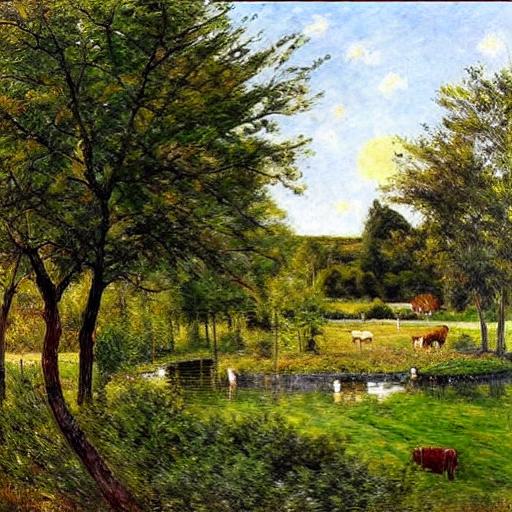}
        \parbox[t]{\txtw}{\raggedright
        \textbf{Context}: A lush countryside scene with trees, a small pond, a house, and cows grazing.\\
        \textbf{Art-Style}: Impressionism.\\
        \textbf{Unity} is achieved by the consistent theme of a rural landscape, with all elements fitting together seamlessly.}
    \end{minipage}\hfill
    \begin{minipage}[t]{\colw}
        \centering
        \includegraphics[width=\imgw]{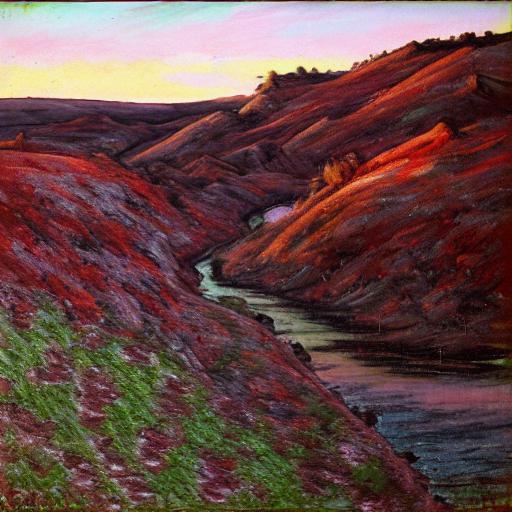}
        \parbox[t]{\txtw}{\raggedright
        \textbf{Context}: Landscape with rolling hills, a winding river, and ruins of a castle under a colorful sky.\\
        \textbf{Art-Style}: Impressionism.\\
        \textbf{Contrast} is created by the dark shadows of the hills and the bright colors of the sky.}
    \end{minipage}
    \vspace{1em} %
    %
    % --- ROW 2 ---
    \begin{minipage}[t]{\colw}
        \centering
        \includegraphics[width=\imgw]{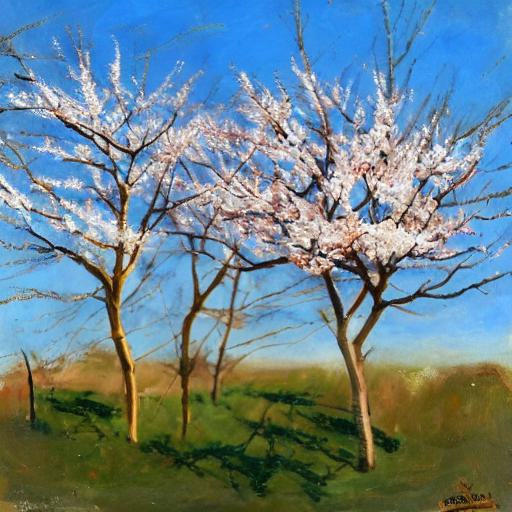}
        \parbox[t]{\txtw}{\raggedright
        \textbf{Context}: A landscape with blooming trees and a grassy foreground under a blue sky.\\
        \textbf{Art-Style}: Impressionism.\\
        \textbf{Emphasis} is placed on the blooming trees, which are the focal point due to their central placement.}
    \end{minipage}\hfill
    \begin{minipage}[t]{\colw}
        \centering
        \includegraphics[width=\imgw]{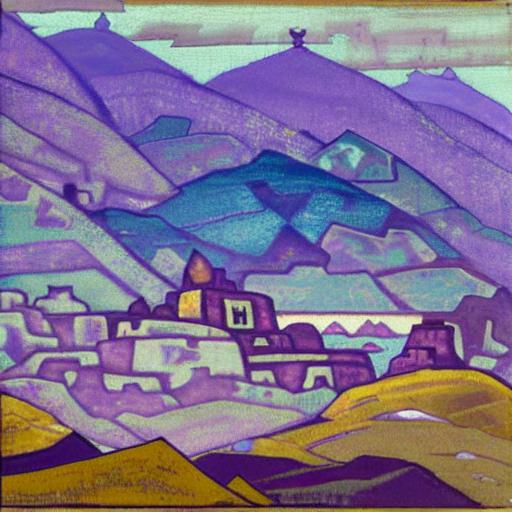}
        \parbox[t]{\txtw}{\raggedright
        \textbf{Context}: Mountainous landscape with layered hills and silhouettes of structures under a cloudy sky.\\
        \textbf{Art-Style}: Symbolism.\\
        \textbf{Proportion} is used to create a sense of depth, with larger structures in the foreground and smaller mountains.}
    \end{minipage}\hfill
    \begin{minipage}[t]{\colw}
        \centering
        \includegraphics[width=\imgw]{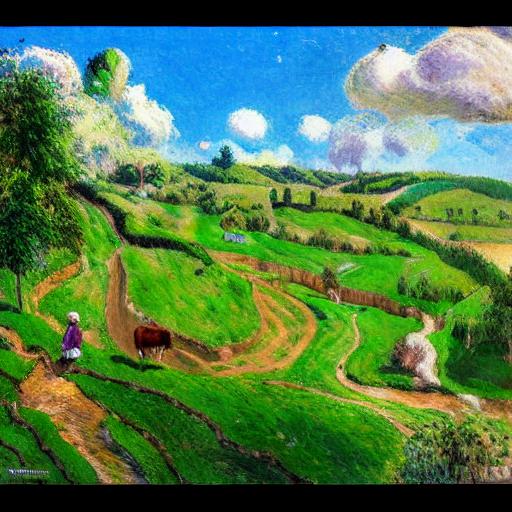}
        \parbox[t]{\txtw}{\raggedright
        \textbf{Context}: A pastoral scene with a woman, cows grazing, and a winding path through a green landscape.\\
        \textbf{Art-Style}: Impressionism.\\
        \textbf{Movement} is suggested by the winding path leading the viewer's eye through the landscape.}
    \end{minipage}\hfill
    \begin{minipage}[t]{\colw}
        \centering
        \includegraphics[width=\imgw]{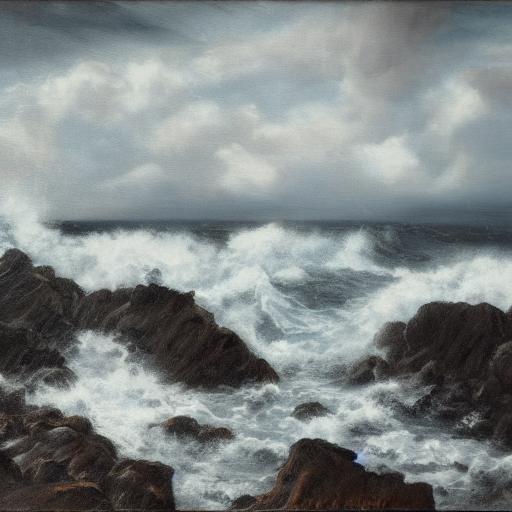}
        \parbox[t]{\txtw}{\raggedright
        \textbf{Context}: A stormy seascape with crashing waves and rocks on the shore under a cloudy sky.\\
        \textbf{Art-Style}: Realism.\\
        \textbf{Rhythm} is present in the repetitive patterns of the waves and the clouds, creating continuity.}
    \end{minipage}\hfill
    \begin{minipage}[t]{\colw}
        \centering
        \includegraphics[width=\imgw]{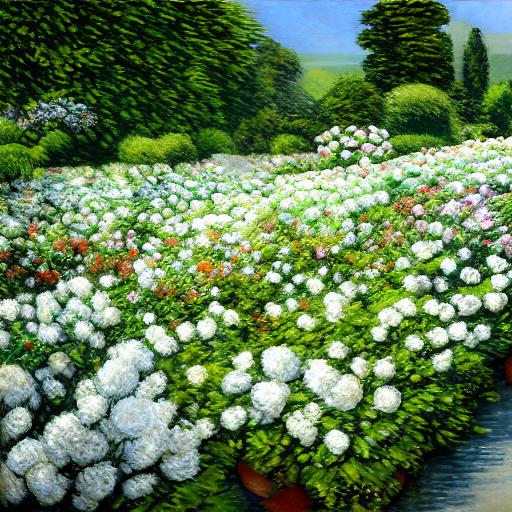}
        \parbox[t]{\txtw}{\raggedright
        \textbf{Context}: A lush garden with abundant white flowers and green foliage, with water in the background.\\
        \textbf{Art-Style}: Impressionism.\\
        \textbf{Pattern} is present in the consistent arrangement of the flowers and leaves, adding a sense of order.}
    \end{minipage}
  }
  \vspace{-1em}
  \caption{Samples generated with our ArtDapter model, showcasing its ability to adhere to the respective context, art-style and compositional controls (e.g., Emphasis, Variety) across the 10 Principles of Art (PoA).}
  \label{fig:teaser}
\end{teaserfigure}

% \received{13 February 2026}
% \received[revised]{? March 2026}
% \received[accepted]{? June 2026}

\maketitle

% Topic sentences important in intro
% Intro: 3/4 page to 1 page

\begin{table*}[ht]
    \centering
    \caption{\textbf{The Vocabulary of Composition:} We operationalize aesthetics using these 10 Principles of Art (PoA). Each principle serves as a distinct control dimension in our framework.}
    \label{tab:poa_definitions}
    \resizebox{\textwidth}{!}{%
    \begin{tabular}{@{}l|l|l@{}}
    \toprule
    \textbf{Principle} & \textbf{Definition} & \textbf{Visual effect \& function} \\
    \midrule
    \textbf{Balance} & Distribution of visual weight (symmetric, asymmetric, or radial). & Brings equilibrium and stability; prevents the image from appearing ``heavy'' on one side. \\
    \textbf{Harmony} & Cohesiveness of elements appearing coordinated. & Creates a sense of order and organization; prevent haphazardness. \\
    \textbf{Variety} & Diversity and complexity of visual elements. & Adds interest, excitement, and dynamism by inviting visual exploration of different areas; prevents monotony. \\
    \textbf{Unity} & Coherence or oneness of elements in serving the context, theme, or message. & Ensures elements belong together semantically and visually; brings meaning and completeness to the work. \\
    \textbf{Contrast} & Use of opposing visual elements (e.g. light/dark, big/small) to highlight differences. & Adds drama and focus by creating strong distinctions that direct attention to key areas. \\
    \textbf{Emphasis} & Creation of focal points that draw attention to important parts of the artwork. & Guides the viewer to the main subject(s), strengthening communication; prevents a directionless viewing experience. \\
    \textbf{Proportion} & Relationship between elements based on their relative sizes. & Establishes perceived depth, realism and importance of elements through scale; cues relationships between objects. \\
    \textbf{Movement} & Suggestion of motion or directional and sequential flow. & Guides the gaze sequentially through the visual narrative; adds dynamism, narrative flow, and sometimes tension. \\
    \textbf{Rhythm} & Visual tempo created through repetition, progression, or alternation of elements. & Creates continuity and flow (calm or energetic); strings distinct parts into a connected sequence. \\
    \textbf{Pattern} & Consistent, organized repetition of elements, often in decorative arrangements. & Adds texture and structure; creates predictability and decorative appeal. \\
    \bottomrule
    \end{tabular}%
    }
    \vspace{-6pt}
\end{table*}

\section{Introduction}
\label{sec:intro}

Diffusion Models (DMs) have driven remarkable advances in text-to-image (T2I) generation, producing high-fidelity outputs~\cite{DBLP:conf/nips/HoJA20,Podell2023SDXLIL}. While the community has achieved significant success in \textit{semantic alignment}, which ensures the model generates \textit{what} is described in the prompt (e.g., ``a dog''), the challenge of \textit{aesthetic alignment}, which controls \textit{how} the image is composed, remains largely unsolved.

In this work, we turn compositional art theory into a practical control interface for T2I diffusion: we (i) define PoA-based constraints, (ii) build a large-scale PoA-annotated dataset, and (iii) learn a lightweight adapter that maps those constraints into diffusion guidance.

Existing T2I approaches on improving the visual appeal of outputs typically sidestep the complexity of aesthetics, treating it as a universal property inferred from large-scale user preferences. Methods often rely on ``prompt engineering'' with vague keywords (e.g., ``masterpiece'', ``trending on artstation'') or fine-tuning on curated datasets like LAION-Aesthetics~\cite{6247954,DBLP:conf/eccv/DattaJLW06}. While effective for improving general visual appeal, these methods suffer from a fundamental limitation: they conflate ``aesthetics'' with a specific, popular visual style, ignoring the complexities of artistic intent.

\textbf{Addressing visual literacy.}
Current T2I interfaces provide strong control over \emph{entities} and sometimes \emph{relations}, but they offer no explicit handle for global compositional attributes. The core issue lies in a lack of \textit{visual literacy} in current T2I control mechanisms. As a result, users must rely on vague aesthetic keywords or trial-and-error prompt edits that do not correspond to stable, interpretable controls. Efforts to decompose image aesthetics into interpretable dimensions are not new. Prior work~\cite{10.24963/ijcai.2024/849, LiuBridgingCognitiveGap, jin2024apddv2} on aesthetics assessment has proposed various such frameworks. However, they conflate subjective aspects (e.g., emotion, mood, creativity) with objective ones (e.g., perception, layout, composition). More importantly, they were all designed to produce a single aesthetic score, a goal that presupposes the universality of aesthetic judgment. This runs contrary to our aim of decomposing aesthetics into user-specifiable dimensions. Moreover, these aesthetics frameworks were not designed with image generation in mind. What we seek is an objective visual literacy — one tailored to generation rather than scoring. Just as grammar structures words into meaningful and beautiful sentences, visual creation needs a grammar of \textit{compositional rules} that organizes visual elements into coherent and appealing imagery.

\textbf{Principles of Art as Control Specifications.} To address this missing interface, we turn to the Principles of Art (PoA), the standard pedagogical framework used in art education to analyze visual composition~\cite{UCBdesignfundamentals,Zhao2014ExploringPF}. PoA decomposes the complex concept of ``aesthetics'' into disentangled, controllable dimensions: balance, harmony, variety, unity, contrast, emphasis, proportion, movement, rhythm, and pattern. We define each of these principles in~\cref{tab:poa_definitions}. By formalizing aesthetics through PoA, we can represent composition as explicit PoA constraints and condition diffusion on them, turning aesthetic control from a black-box optimization problem into a user-specifiable one.

We introduce \textit{Aesthetic Alignment}, a task that seeks to align generated outputs with explicit PoA constraints. We construct \textbf{CompArt}, a large-scale dataset of artworks annotated with PoA analyses, enabling models to learn these compositional concepts. We then propose \textbf{ArtDapter}, a lightweight adapter that disentangles these PoA attributes, allowing users to steer the diffusion process along specific aesthetic dimensions.

Our contributions are:
\begin{enumerate}[leftmargin=*]
    \item \textbf{Task:} We formulate \textit{Aesthetic Alignment} for T2I generation as adherence to explicit compositional constraints specified via the Principles of Art (PoA).
    \item \textbf{Data:} We introduce \textbf{CompArt} (80,032 artworks) with automatically generated captions and structured PoA analyses (with prominence levels) produced by a multimodal LLM, enabling supervised learning of PoA-conditioned generation. The dataset is accessible at \hyperlink{https://huggingface.co/datasets/thejinzhe/CompArt}{https://huggingface.co/datasets/thejinzhe/CompArt}.
    \item \textbf{Method:} We propose \textbf{ArtDapter}, a parameter-efficient adapter that conditions a frozen diffusion model on caption, style, and PoA controls in a disentangled manner. We release our code for public access at \hyperlink{https://github.com/jin-zhe/ArtDapter}{https://github.com/jin-zhe/ArtDapter}.
    \item \textbf{Evaluation:} We present a two-level evaluation protocol at the principle and composition levels, combining LLM-based consistency checking with an external preference model as an additional signal.
\end{enumerate}

\noindent\textbf{Positioning.} We study composition-level control rather than style transfer or local layout constraints. Unlike personalized stylization methods that bind a concept/style token to a look, our controls are interpretable compositional principles (e.g., balance, emphasis) that should generalize across styles. Unlike layout-controlled generation that specifies explicit spatial boxes or relations, PoA constraints are semi-global and describe how visual elements distribute and interact across the image. Our dataset and adapter are designed specifically for this intermediate level of control.

% Dataset/method (incl analysis): 3-3.5 pages. Leading paragraph to introduce subsections/studies.
\section{CompArt dataset}

\input{fig/compart_principle_wise}

To facilitate the study of T2I aesthetic alignment via PoA, we construct \textbf{CompArt}, a large-scale art dataset with PoA annotations. CompArt extends the WikiArt dataset~\cite{Saleh2015LargescaleCO}, which comprises 80,032 artworks scraped in 2015 from the public visual-art encyclopedia WikiArt\footnote{\url{https://www.wikiart.org/}}. The artworks cover 1,119 artists across 27 styles, and each is labeled with artist, style, and genre. We extend these labels with content captions (averaging 19.1 words) and PoA annotations (averaging 25.5 words). Unlike~\citet{bai2021explain}, who generate multi-topic artwork descriptions under the framework of content, form, and context, we focus on compositional analysis via PoA, independent of artwork context. From their perspective, our PoA annotations can be seen as a principled and fine-grained decomposition of ``form''.

\begin{figure}[t]
    \centering
    \newlength{\compartcolsep}
    \newlength{\compartcolwidth}
    \setlength{\compartcolsep}{1pt}

    \setlength{\compartcolwidth}{\dimexpr(\linewidth-4\compartcolsep-2pt)/5\relax}

    \captionsetup[subfigure]{
        font={tiny,bf},
        labelfont=bf,
        labelformat=parens,
        labelsep=space,
        justification=centering,
        singlelinecheck=false,
        skip=2pt
    }

    \newcommand{\compartgalleryitem}[4]{%
        \subcaptionbox{#2\label{fig:compart-gallery-#4}}[\linewidth][c]{%
            \includegraphics[width=\linewidth]{#1}%
        }%
        \par\vspace{0.5pt}
        {\tiny\raggedright #3\par}
        \vspace{2pt}
    }

    \makebox[\linewidth][c]{%
    \begin{minipage}[t]{\compartcolwidth}
        \vspace{0pt}
        \compartgalleryitem
        {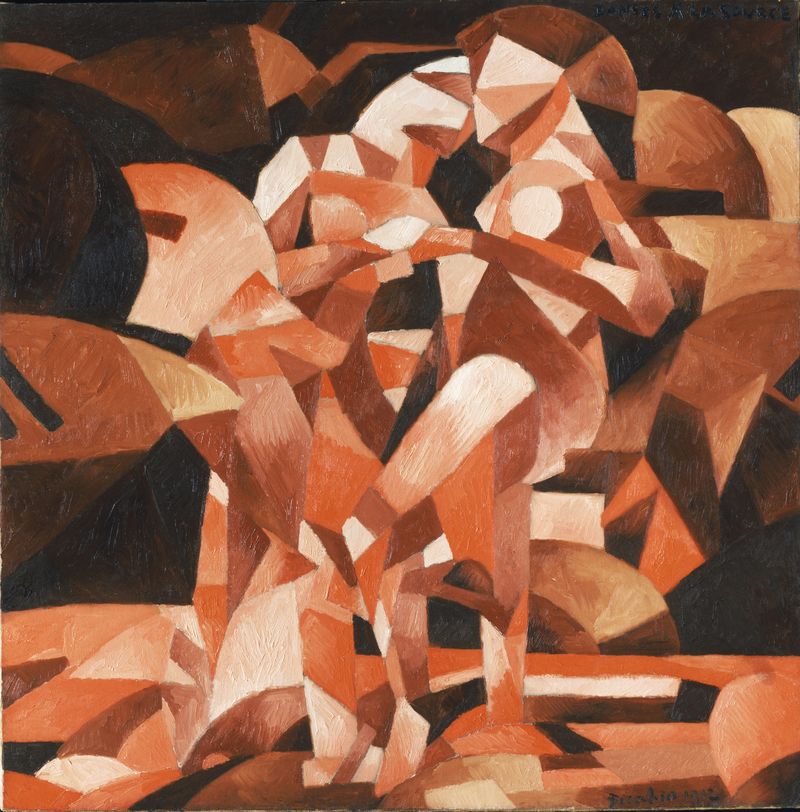}
        {Caption}
        {Two abstract human figures composed of geometric shapes in warm tones.}
        {caption}

        \compartgalleryitem
        {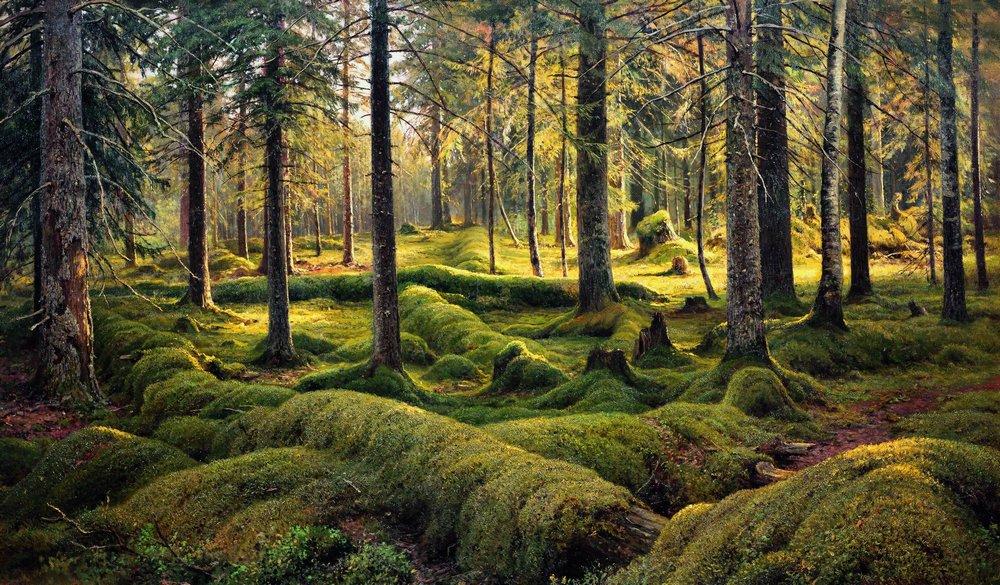}
        {Harmony}
        {Harmony is achieved through the consistent use of natural elements such as trees, moss, and sunlight, which create a cohesive and organized composition. The uniformity of the green hues and the repetition of tree trunks contribute to the overall sense of order and cohesiveness.}
        {harmony}

        \compartgalleryitem
        {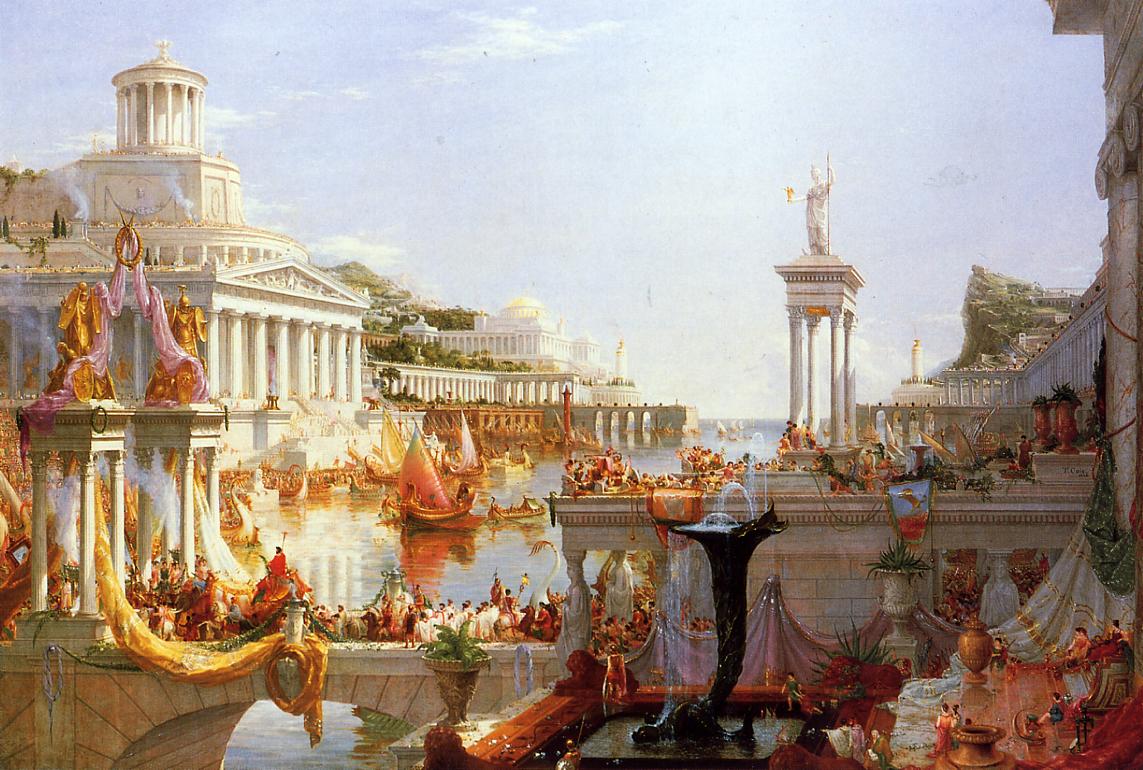}
        {Variety}
        {Variety is present in the diverse activities of the people, the different architectural structures, and the various boats in the harbor, adding interest and complexity to the composition.}
        {variety}
    \end{minipage}%
    \hspace{\compartcolsep}%
    \begin{minipage}[t]{\compartcolwidth}
        \vspace{0pt}
        \compartgalleryitem
        {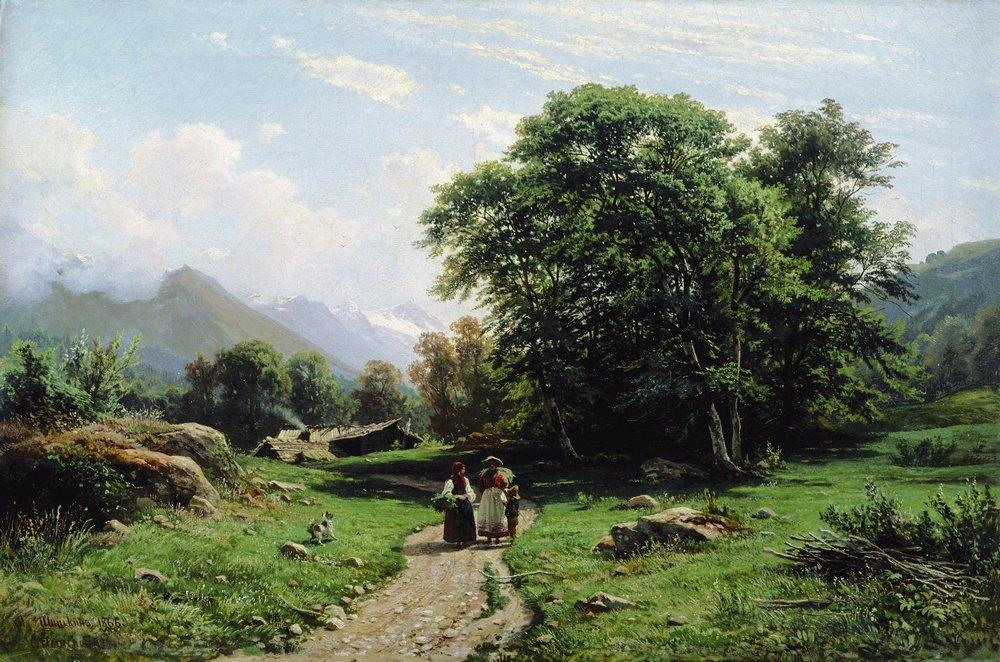}
        {Asymmetric Balance}
        {Asymmetric balance is evident in the composition, with the large tree on the right balanced by the distant mountains and smaller trees on the left, creating a sense of stability and natural equilibrium.}
        {asymmetric-balance}

        \compartgalleryitem
        {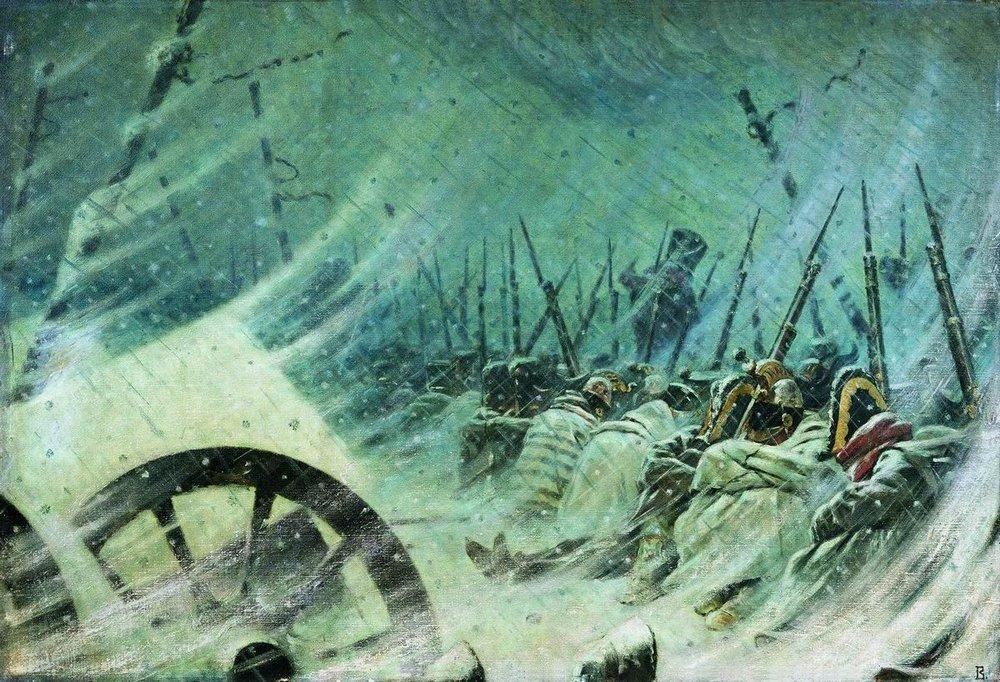}
        {Movement}
        {Movement is suggested by the swirling snow and the diagonal lines created by the soldiers' rifles, guiding the viewer's eye across the composition and conveying the dynamic and chaotic nature of the scene.}
        {movement}

        \compartgalleryitem
        {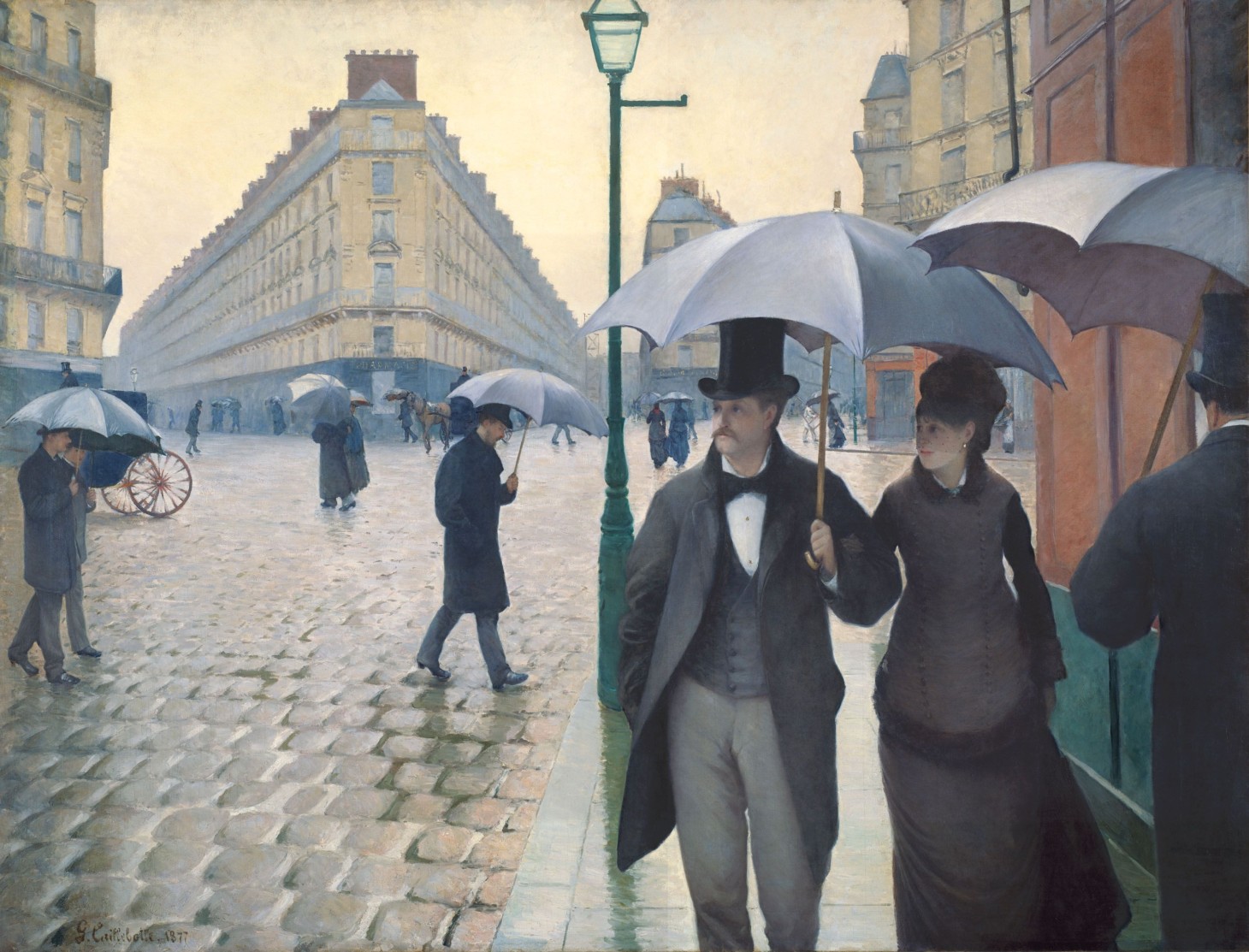}
        {Unity}
        {Unity is achieved by the common theme of rain and umbrellas, which ties all the elements together and creates a coherent scene in the composition.}
        {unity}
    \end{minipage}%
    \hspace{\compartcolsep}%
    \begin{minipage}[t]{\compartcolwidth}
        \vspace{0pt}
        \compartgalleryitem
        {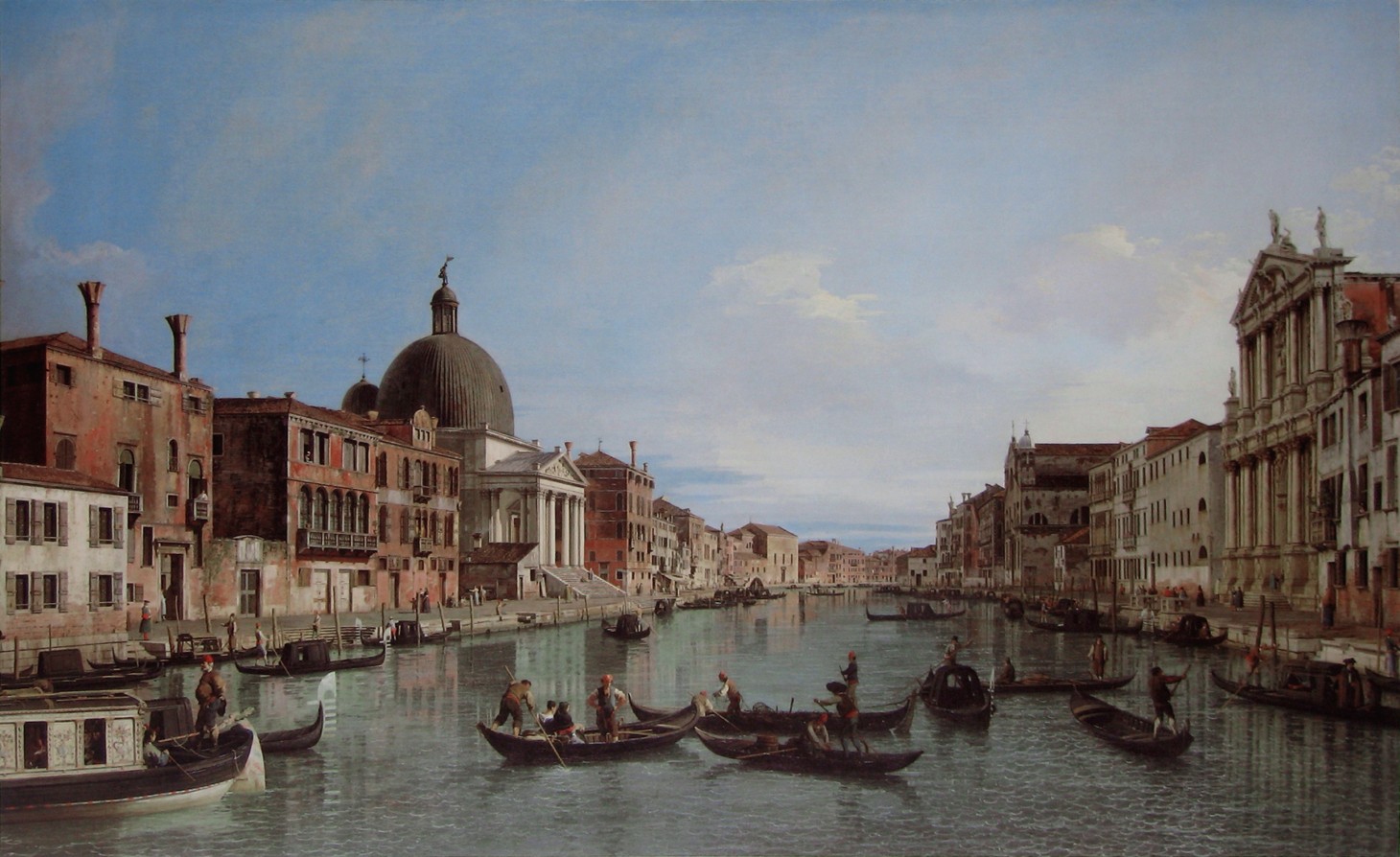}
        {Symmetric Balance}
        {Symmetric balance is evident in the composition, with the canal serving as a central axis and the buildings on both sides mirroring each other in terms of height and architectural style, creating a sense of equilibrium.}
        {symmetric-balance}

        \compartgalleryitem
        {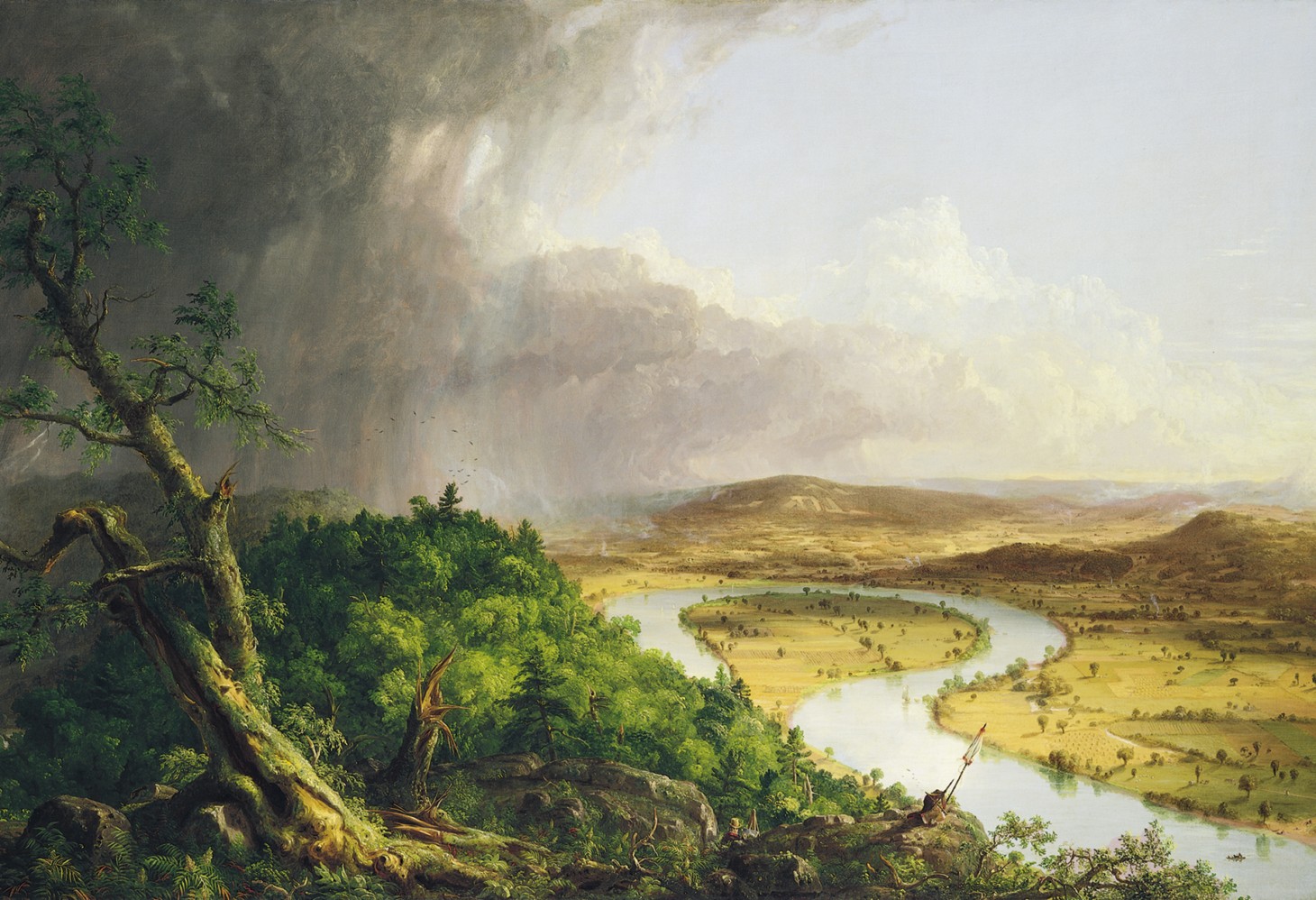}
        {Contrast}
        {Contrast is strong between the dark, stormy sky on the left and the bright, clear sky on the right, highlighting the dramatic change in weather and drawing attention to the central river.}
        {contrast}

        \compartgalleryitem
        {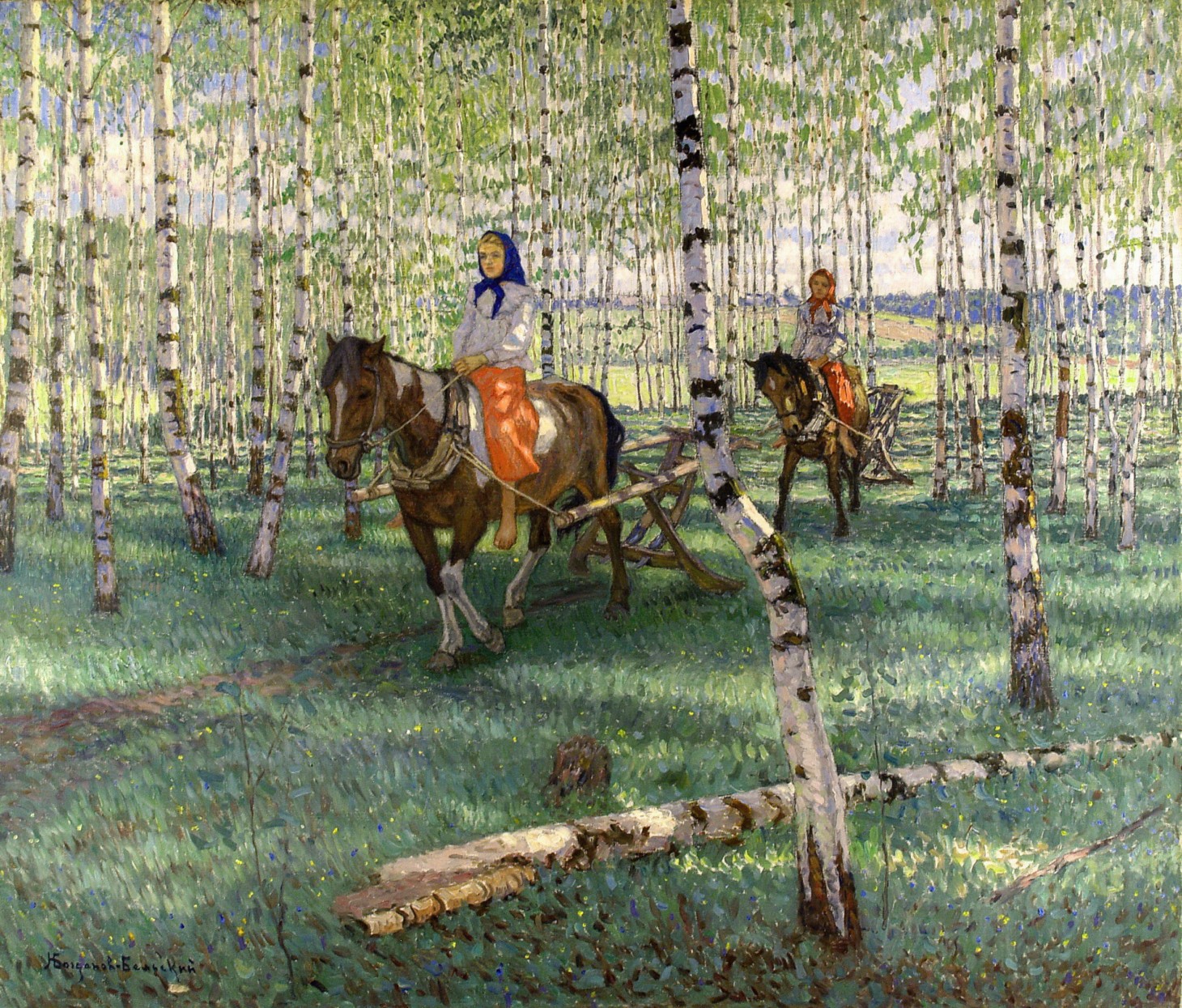}
        {Rhythm}
        {Rhythm is created by the repetitive vertical lines of the birch trees, leading the viewer's eye across the composition in a steady and calming manner.}
        {rhythm}
    \end{minipage}%
    \hspace{\compartcolsep}%
    \begin{minipage}[t]{\compartcolwidth}
        \vspace{0pt}
        \compartgalleryitem
        {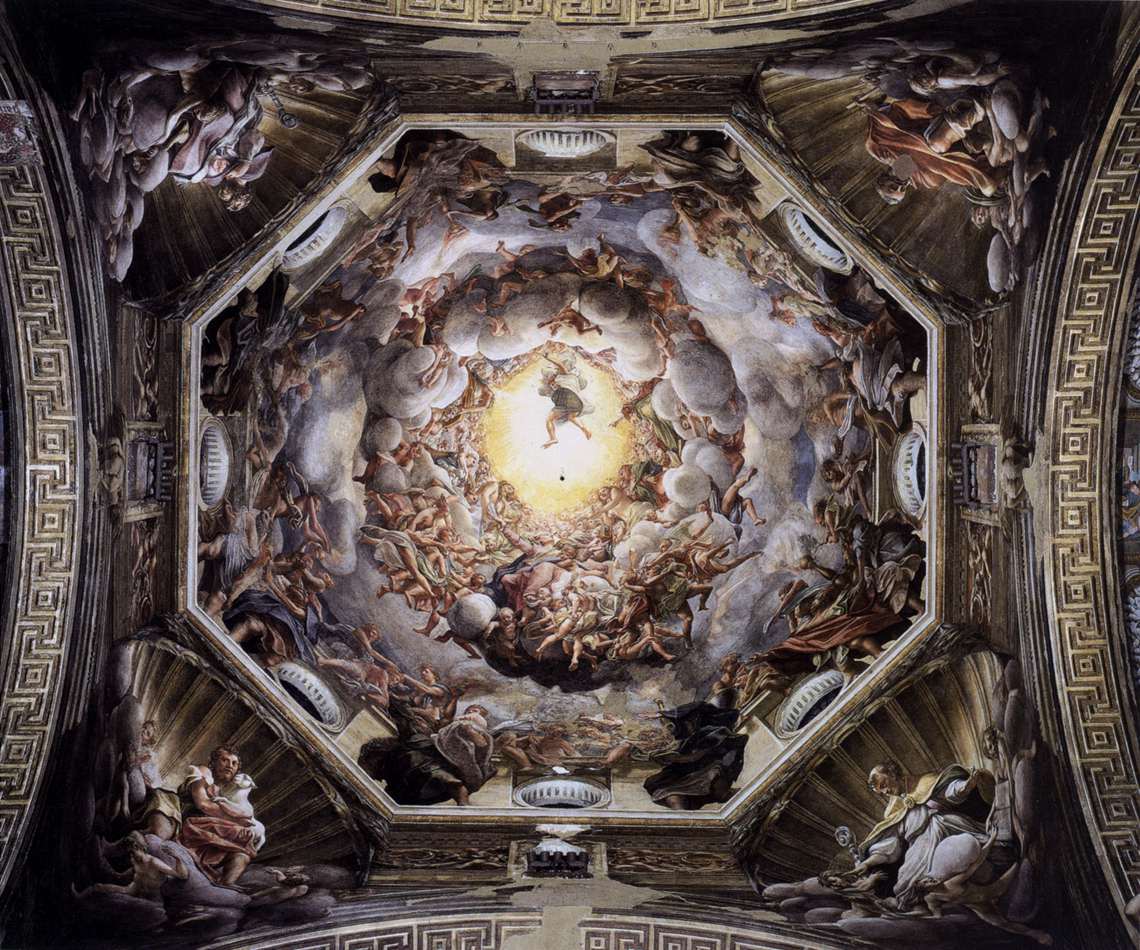}
        {Radial Balance}
        {Radial balance is evident in the composition, with figures and architectural elements radiating outwards from the central light source, creating a sense of harmony and stability.}
        {radial-balance}

        \compartgalleryitem
        {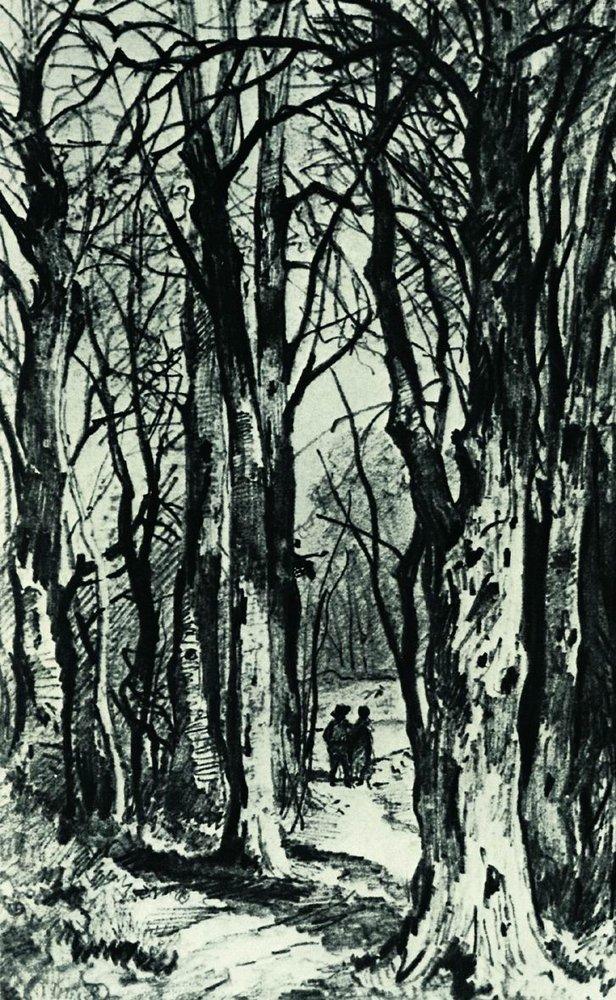}
        {Proportion}
        {Proportion is evident in the towering height of the trees compared to the small figures, emphasizing the grandeur of nature and the insignificance of humans within it.}
        {proportion}
    \end{minipage}%
    \hspace{\compartcolsep}%
    \begin{minipage}[t]{\compartcolwidth}
        \vspace{0pt}
        \compartgalleryitem
        {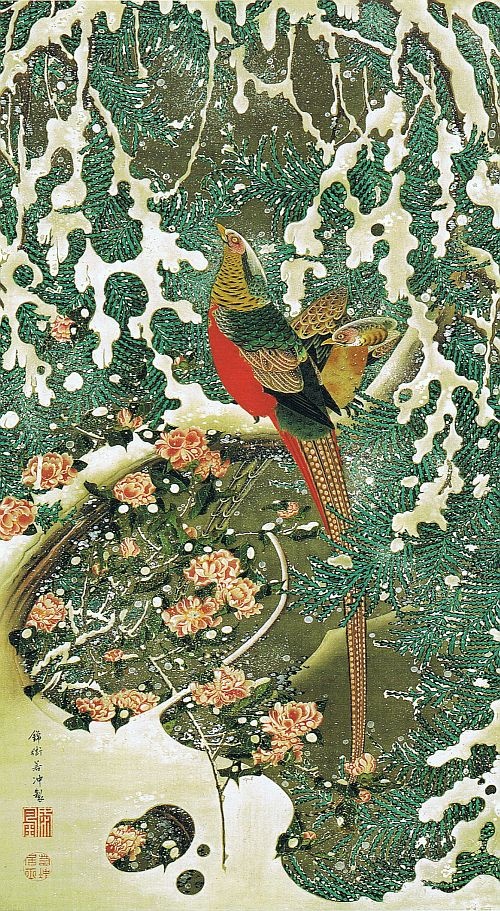}
        {Pattern}
        {Pattern is evident in the consistent and organized repetition of leaves and flowers, which adds texture and visual appeal to the composition.}
        {pattern}

        \compartgalleryitem
        {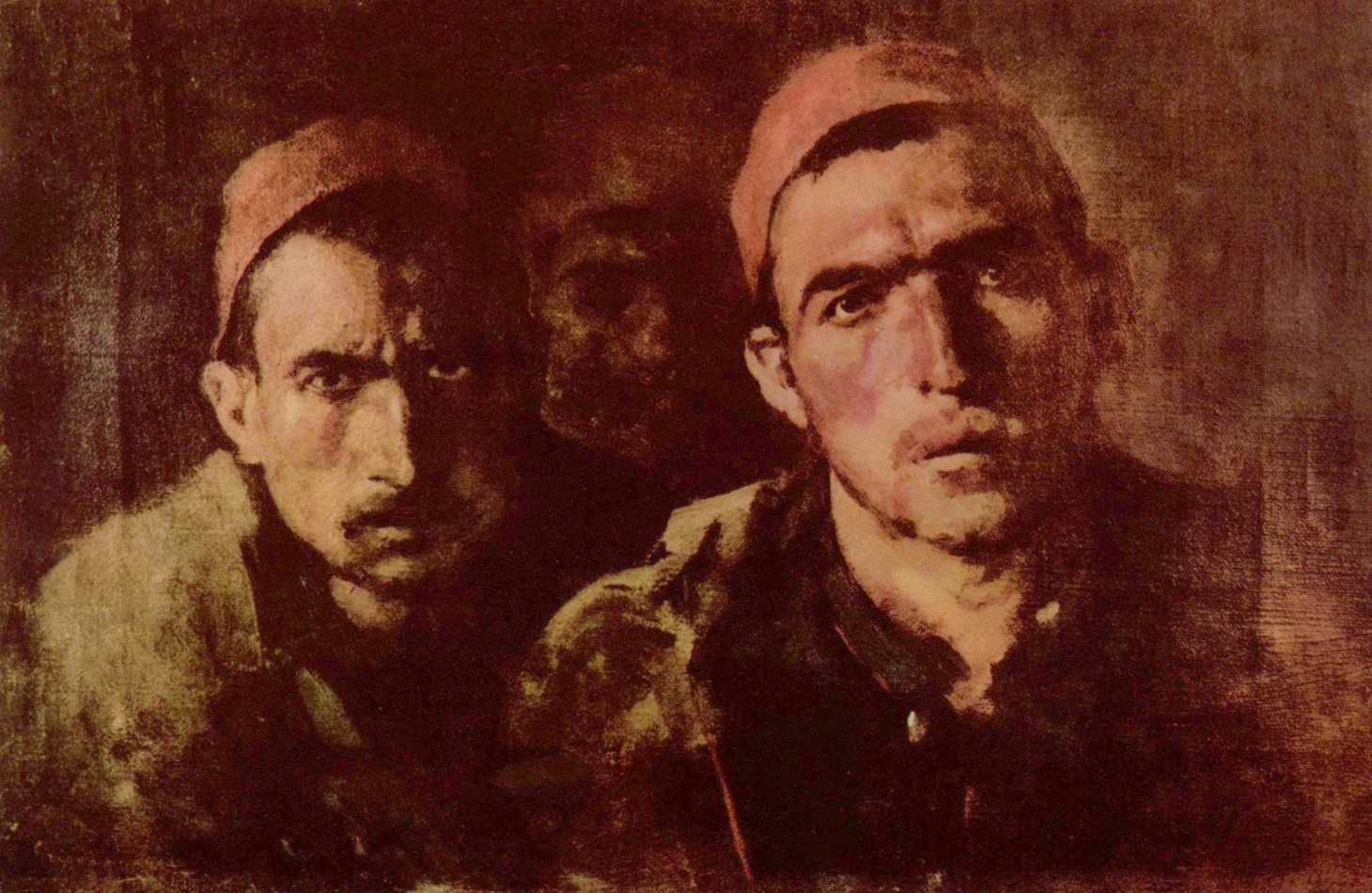}
        {Emphasis}
        {Emphasis is placed on the men's faces, particularly their eyes, through the use of lighting and positioning. Their intense expressions are the focal point of the composition, drawing the viewer's attention immediately.}
        {emphasis}
    \end{minipage}%
    }
    \caption{CompArt annotation examples. We provide one example per annotation type — the content caption and each of the 10 principles of art — with separate examples for Balance's asymmetric, symmetric, and radial senses.}
    \squeezegap
    \Description{A gallery of artwork examples illustrating captions and each type of principles of art annotations, covering balance, harmony, variety, emphasis, movement, unity, contrast, rhythm, pattern, and proportion.}
    \label{fig:compart-gallery}
\end{figure}

In total, CompArt contains \textbf{637{,}573} PoA annotations totaling \textbf{17{,}800{,}136} words. \cref{fig:compart-gallery} shows an example of each annotation type. We partition CompArt into train and test splits of 79,032 and 1,000 images respectively. \cref{fig:PoA_breakdown} shows a per-principle breakdown of annotation counts.

In the following sections, we detail the annotation method of CompArt and present some important analyses.

\subsection{Annotation}
Annotating a dataset of this scale with PoA using human experts is infeasible in both cost and time. Such an undertaking has only recently become feasible with the emergent capabilities of multi-modal large language models (MLLMs). We therefore employ OpenAI's GPT-4o ({\footnotesize \texttt{gpt-4o-2024-05-13}})~\footnote{\url{https://openai.com/index/hello-gpt-4o/}}, instructing it to act as an expert art analyst. This choice is motivated by prior evidence that GPT-4–class models can match or exceed human annotators at creative visual interpretation of visual stimuli~\cite{doi:10.1080/10447318.2024.2345430}. Using an MLLM enables a consistent rubric at scale, which is difficult to achieve with distributed annotators.

\textbf{Annotation Constraints.} To produce high-quality, structured image–PoA pairs, we enforced strict annotation protocols:
\begin{itemize}[leftmargin=*]
    \item \textbf{Caption objectivity:} Captions must be concise and purely describe image content, avoiding subjective praise (e.g., ``beautiful'') and any meta-reference to the image as an artwork or to its medium (e.g., ``oil painting''). This leverages the generalization of the pre-trained T2I backbone and maximizes the ArtDapter's sensitivity to the artistic controls.
    \item \textbf{Prominence scaling:} For each PoA, the annotator assigns a prominence level from \{``weak'', ``mild'', ``moderate'', ``strong''\}. Detailed analysis is optional for principles rated ``weak'' and required otherwise, avoiding padded entries on marginal principles while retaining broad coverage.
    \item \textbf{Structured analysis:} Each non-optional analysis must concisely identify \textit{where} in the composition the principle is evident, \textit{which} visual elements are involved, \textit{how} the principle is achieved, and \textit{what} effect it produces (e.g., ``Proportion is used to create a sense of depth, with larger structures in the foreground and smaller mountains.''). For the principle of Balance, its stated sense must be one of \{``symmetric'', ``asymmetric'', ``radial''\}.
    \item \textbf{Reporting consistency:} The first sentence of every analysis must take the principle itself as its subject, and the image must be referred to only as ``the composition'' — never by style or medium — so that art-style descriptors do not contaminate the principle-level analysis.
\end{itemize}

\textbf{How to interpret CompArt labels.}
We treat the PoA analyses as structured \emph{pseudo-labels}: they provide a consistent, scalable description of compositional cues that is useful for learning controllable generation. They are not a substitute for expert consensus, and we therefore design evaluation to (i) check consistency with the same labeling scheme and (ii) include an external preference model as an additional signal (Sec.~\cref{sec:evaluation_scheme}).

\begin{table}[t]
    \caption{Predictive performance of GTP-4o on the art-style of CompArt images, presented in Top[1,2,3] accuracies. The proportion (Prop) of each style in the dataset is indicated in the first column of each table section.}
    \label{tab:style_predictions}
    \tiny
    \centering
    \setlength{\tabcolsep}{0.7pt}
    \begin{tabular}{@{}%
        c l c c c | c l c c c | c l c c c
        @{}}
        \toprule
        \makecell{\textbf{Prop}\\(\%)} &
        \makecell{\textbf{Art-Style}\\\strut} &
        \makecell{\textbf{Top1}\\(\%)} &
        \makecell{\textbf{Top2}\\(\%)} &
        \makecell{\textbf{Top3}\\(\%)} &
        \makecell{\textbf{Prop}\\(\%)} &
        \makecell{\textbf{Art-Style}\\\strut} &
        \makecell{\textbf{Top1}\\(\%)} &
        \makecell{\textbf{Top2}\\(\%)} &
        \makecell{\textbf{Top3}\\(\%)} &
        \makecell{\textbf{Prop}\\(\%)} &
        \makecell{\textbf{Art-Style}\\\strut} &
        \makecell{\textbf{Top1}\\(\%)} &
        \makecell{\textbf{Top2}\\(\%)} &
        \makecell{\textbf{Top3}\\(\%)} \\
        \midrule
        % ---- Row 1
        0.12 & \makecell[l]{Action\\painting} & 39.8 & 85.71 & 90.82 &
        1.60 & \makecell[l]{Mannerism\\(Late\\Renaissance)} & 50.0 & 69.41 & 83.1 &
        3.36 & \makecell[l]{Abstract\\Expressionism} & 82.05 & 96.99 & 98.92 \\
        \hline
        % ---- Row 2
        0.14 & \makecell[l]{Analytical\\Cubism} & 94.55 & 98.18 & 98.18 &
        1.67 & \makecell[l]{High\\Renaissance} & 76.94 & 92.24 & 95.0 &
        5.29 & Symbolism & 48.9 & 62.15 & 74.51 \\
        \hline
        % ---- Row 3
        0.27 & \makecell[l]{Synthetic\\Cubism} & 27.31 & 92.13 & 99.07 &
        1.74 & \makecell[l]{Early\\Renaissance} & 77.9 & 96.04 & 98.13 &
        5.30 & Baroque & 65.72 & 70.39 & 82.78 \\
        \hline
        % ---- Row 4
        0.39 & \makecell[l]{New Realism} & 0.0 & 0.64 & 20.7 &
        1.85 & \makecell[l]{Pop Art} & 51.99 & 63.99 & 70.73 &
        5.34 & \makecell[l]{Art Nouveau\\(Modern)} & 36.77 & 44.87 & 52.13 \\
        \hline
        % ---- Row 5
        0.60 & \makecell[l]{Contemporary\\Realism} & 41.58 & 72.97 & 86.69 &
        2.00 & \makecell[l]{Color Field\\Painting} & 31.9 & 67.85 & 89.95 &
        8.02 & \makecell[l]{Post-\\Impressionism} & 49.47 & 76.21 & 85.87 \\
        \hline
        % ---- Row 6
        0.64 & Pointillism\makecell[l]{\\\strut} & 82.42 & 84.77 & 89.65 &
        2.52 & Cubism & 62.67 & 70.45 & 74.5 &
        8.29 & Expressionism & 40.72 & 62.55 & 70.75 \\
        \hline
        % ---- Row 7
        0.93 & Fauvism\makecell[l]{\\\strut} & 46.39 & 57.09 & 67.11 &
        2.61 & Rococo & 39.39 & 52.61 & 58.84 &
        8.62 & Romanticism & 34.72 & 83.29 & 92.36 \\
        \hline
        % ---- Row 8
        1.45 & {Ukiyo-e} & 99.74 & 99.74 & 99.74 &
        3.00 & \makecell[l]{Naïve Art\\(Primitivism)} & 51.66 & 59.9 & 78.45 &
        13.40 & Realism & 80.09 & 89.85 & 95.64 \\
        \hline
        % ---- Row 9
        1.57 & Minimalism & 85.49 & 94.58 & 98.72 &
        3.19 & \makecell[l]{Northern\\Renaissance} & 84.59 & 87.02 & 91.88 &
        16.08 & Impressionism & 68.0 & 84.46 & 89.75 \\
        \bottomrule
        \multicolumn{10}{c|}{} & 100.0 & All & 58.9 & 76.3 & 84.1 \\
    \end{tabular}
    \squeezegap
\end{table}

\subsection{Quantitative analysis}

\textbf{Annotator proficiency}. Since GPT-4o serves as the annotator of CompArt, we investigate its artistic proficiency. Although MLLMs exhibit strong vision-language performance, the artistic domain remains challenging. Even humans without an art background or cultural/historical context struggle to identify abstractions and motifs correctly~\cite{7533051}. We quantitatively assessed GPT-4o's ability to correctly predict the art-style of artworks in the dataset. Here, it is important to note that art-styles are not simply visual stylization, but rather \textit{art movements} from different periods, captured by characteristic techniques, media, motifs, and subjects. Hence, art-style prediction provides a quantitative measure of GPT-4o's \textit{artistic knowledge}, which is crucial in its role as an art analyst. Across 27 art-styles, GPT-4o achieves Top[1,2,3] accuracies of 58.9\%, 76.3\%, and 84.1\% respectively. A full breakdown of a per-style Top[1,2,3] accuracy is reported in \cref{tab:style_predictions}.

\textbf{Linguistic analysis.} Following \citet{Achlioptas2021ArtEmisAL}, we report CompArt's linguistic richness and diversity via word count and parts-of-speech (nouns, pronouns, adjectives, verbs, adpositions) in \cref{tab:richness_diversity}, comparing against existing captioning datasets~\cite{Achlioptas2021ArtEmisAL,Mohamed2022ItIO,DBLP:journals/corr/ChenFLVGDZ15,DBLP:conf/acl/SoricutDSG18,DBLP:journals/tacl/YoungLHH14,DBLP:conf/cvpr/MaoHTCY016}. Treating captions and PoA analyses as homogeneous utterances, CompArt leads in word count (by 24.8 words per annotation) and all PoS categories except pronouns. This indicates higher annotation detail, especially in nouns, adjectives, and adpositions — a signal of richer \textit{visual composition} descriptions. While not always topping diversity scores, CompArt remains competitive, suggesting that its structured analyses retain variety. When captions and PoA analyses are treated as distinct annotation types (lower half of the table), CompArt leads every PoS category by wider margins. Annotations for each principle average 22.5-29.9 words, indicating consistent detail across them.

\begin{table}[t]
    \centering
    \scriptsize
    \setlength{\tabcolsep}{3pt}
    \caption{Comparison of CompArt and prior works on linguistic richness (avg. word tokens per annotation) and diversity (avg. \textit{unique} tokens per image). The top-half table treats captions and 10 PoA analyses as \textit{homogeneous} annotations, ignoring content differences. The bottom-half treats them as \textit{distinct} annotation types.}
    \begin{tabular}{@{} l c cc cc cc cc cc @{}}
        \toprule
        \multicolumn{1}{c}{\textbf{Corpus}} & \textbf{Words} & \multicolumn{2}{c}{\textbf{NOUN}} & \multicolumn{2}{c}{\textbf{PRON}} & \multicolumn{2}{c}{\textbf{ADJ}} & \multicolumn{2}{c}{\textbf{ADP}} & \multicolumn{2}{c}{\textbf{VERB}} \\
        \cmidrule(lr){2-2} \cmidrule(lr){3-4} \cmidrule(lr){5-6} \cmidrule(lr){7-8} \cmidrule(lr){9-10} \cmidrule(lr){11-12}
         & Rch. & Rch. & Div. & Rch. & Div. & Rch. & Div. & Rch. & Div. & Rch. & Div. \\  
        \midrule
        ArtEmis V2~\cite{Mohamed2022ItIO} & 15.9 & 3.9 & 3.2 & \textbf{0.9} & 0.5 & 1.6 & 1.4 & 1.9 & 1.1 & 3.2 & 2.3 \\
        ArtEmis~\cite{Achlioptas2021ArtEmisAL} & 15.9 & 4.0 & 3.4 & \textbf{0.9} & \textbf{0.6} & 1.6 & 1.5 & 1.9 & 1.2 & 3.0 & \textbf{2.4} \\
        COCO Captions~\cite{DBLP:journals/corr/ChenFLVGDZ15} & 10.5 & 3.7 & 2.2 & 0.1 & 0.1 & 0.8 & 0.7 & 1.7 & 0.9 & 1.2 & 0.9 \\
        Conceptual Capt.~\cite{DBLP:conf/acl/SoricutDSG18} & 9.6 & 3.8 & 3.8 & 0.2 & 0.2 & 0.9 & 0.9 & 1.6 & \textbf{1.6} & 1.1 & 1.1 \\
        Flickr30k Ent.~\cite{DBLP:journals/tacl/YoungLHH14} & 12.3 & 4.2 & 2.6 & 0.2 & 0.2 & 1.1 & 0.8 & 1.9 & 1.0 & 1.8 & 1.3 \\
        Google Refexp~\cite{DBLP:conf/cvpr/MaoHTCY016} & 8.4 & 3.0 & 2.2 & 0.1 & 0.1 & 1.0 & 0.8 & 1.2 & 0.8 & 0.8 & 0.6 \\
        \textbf{CompArt} & \textbf{24.8} & \textbf{7.7} & \textbf{4.9} & 0.2 & 0.1 & \textbf{3.1} & \textbf{2.4} & \textbf{3.4} & 1.1 & \textbf{3.3} & 2.1 \\
        \hline
        --- Caption & 19.1 & 6.4 & 6.4 & 0.1 & 0.1 & 2.7 & 2.6 & 3.3 & 2.8 & 1.7 & 1.7 \\
        --- PoA Balance & \textbf{29.9} & 8.5 & 8.3 & 0.1 & 0.1 & \textbf{5.2} & \textbf{5.1} & \textbf{4.7} & \textbf{4.1} & 3.4 & 3.4 \\
        --- PoA Harmony & 24.1 & 6.9 & 6.8 & 0.0 & 0.0 & 4.0 & 4.0 & 2.9 & 2.4 & \textbf{4.0} & \textbf{4.0} \\
        --- PoA Variety & 24.6 & 8.3 & 8.2 & 0.1 & 0.1 & 3.4 & 3.4 & 3.0 & 2.7 & 2.8 & 2.8 \\
        --- PoA Unity & 25.0 & 7.5 & 7.5 & 0.0 & 0.0 & 3.2 & 3.2 & 3.3 & 2.9 & 3.1 & 3.1 \\
        --- PoA Contrast & 26.2 & 8.4 & 8.3 & 0.3 & 0.2 & 3.2 & 3.2 & 3.3 & 2.9 & 3.2 & 3.2 \\
        --- PoA Emphasis & 26.0 & 7.8 & 7.8 & \textbf{1.2} & \textbf{1.0} & 2.5 & 2.5 & 3.2 & 2.9 & \textbf{4.0} & 3.9 \\
        --- PoA Proportion & 23.4 & 7.2 & 7.2 & 0.1 & 0.1 & 2.5 & 2.5 & 3.1 & 2.5 & 3.7 & 3.7 \\
        --- PoA Movement & 25.8 & \textbf{8.7} & \textbf{8.6} & 0.1 & 0.1 & 1.0 & 1.0 & 3.8 & 3.3 & 3.9 & 3.9 \\
        --- PoA Rhythm & 24.5 & 8.4 & 8.4 & 0.1 & 0.1 & 1.8 & 1.8 & 3.7 & 2.6 & 3.2 & 3.2 \\
        --- PoA Pattern & 22.5 & 7.2 & 7.2 & 0.0 & 0.0 & 3.3 & 3.3 & 2.6 & 2.2 & 2.6 & 2.6 \\
        \bottomrule
        \multicolumn{12}{l}{
            \makecell[l]{
            Rch. and Div. respectively refer to Richness and Diversity.\\
            NOUN, PRON, ADJ, ADP and VERB respectively refers to the Noun, Pronoun,\\
            Adjective, Adposition and Verb parts-of-speech.
            }
        }
    \end{tabular}
    \label{tab:richness_diversity}
    \squeezegap
\end{table}

\subsection{Qualitative analysis}

\textbf{Annotator proficiency}. In our qualitative assessment of GPT-4o as the annotator, we first note the limitations officially documented by OpenAI.~\footnote{Archived page: \url{https://archive.ph/KM5yP}}. Crucially, it may misinterpret rotated images, miscount objects, and struggle with precise spatial localization. Images are also resized before analysis, so original proportions are not preserved; intentional proportional choices by the artist may therefore be misread.

\begin{figure}[t]
    \tiny
    \def\capw{.05\linewidth}
    \def\imgw{.2\linewidth}
    \def\margin{.02\linewidth}
    \def\txtw{.73\linewidth}
    \def\rowsep{3pt}
    
    % --- HEADER ROW ---
    \begin{minipage}[t]{\capw}
        \hspace{\fill}
    \end{minipage}%
    \begin{minipage}[t]{\imgw}
        \centering \textbf{Image}
    \end{minipage}%
    \hspace{\margin}%
    \begin{minipage}[t]{\txtw}
        \centering \textbf{Response (\texttt{gpt-4o-2024-05-13})}
    \end{minipage}%
    \hfill % Fills the rest of the line to avoid underfull warnings
    \par % End the header paragraph
    \vspace{2pt} \hrule % Add a line and some space

    % --- CONTENT ROW ---

    % --- ROW 1 ---
    \vspace{\rowsep}
    \begin{subfigure}[t]{\linewidth}
        \begin{minipage}[t]{\capw}
            \caption{}
            \label{fig:girl_with_a_pearl_earring}
        \end{minipage}%
        \begin{minipage}[t]{\imgw}
            \vspace{0pt} % Essential for top alignment with images
            \includegraphics[width=\linewidth]{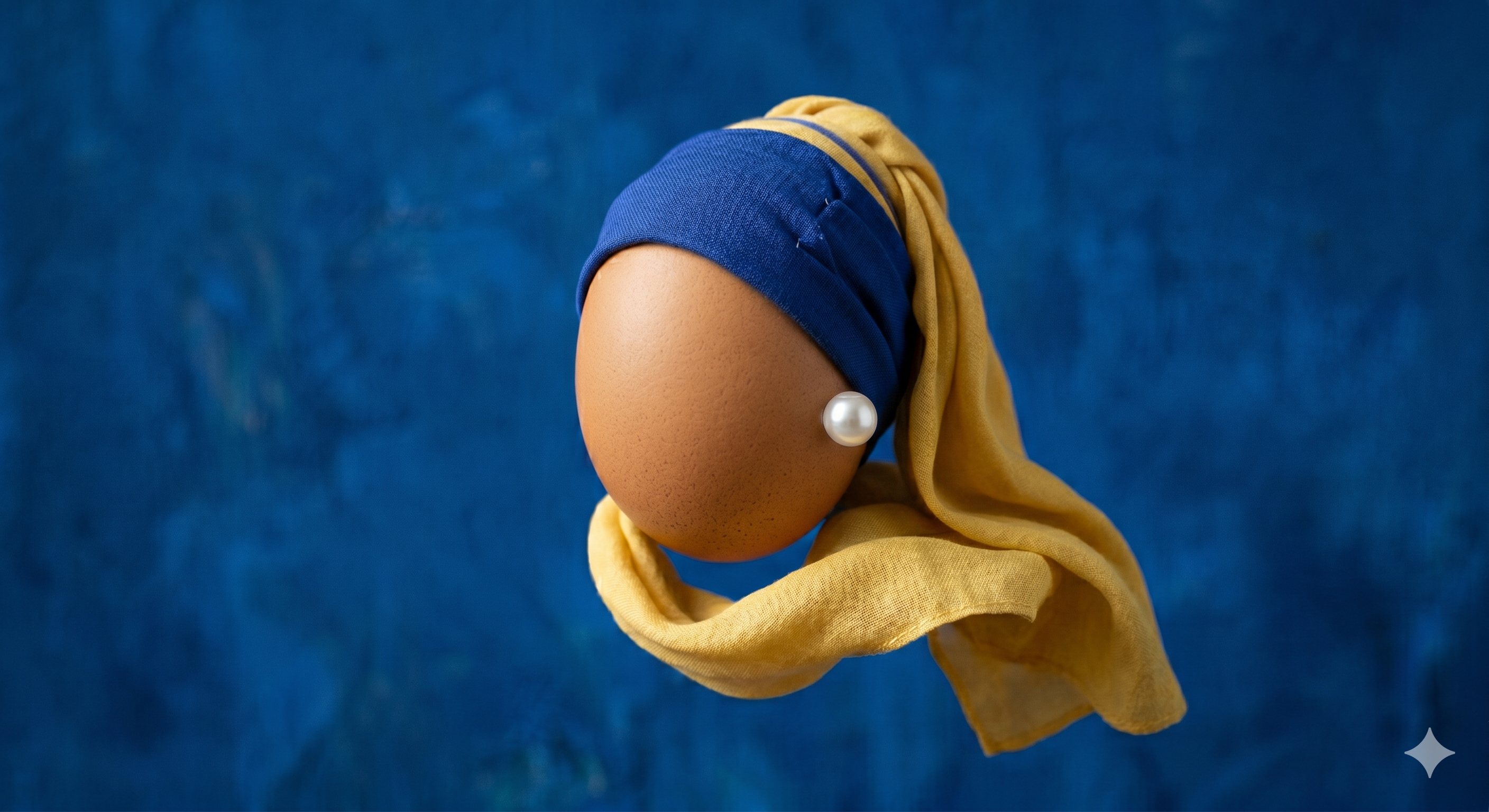}
            {\centering Generated by Google Gemini \par}
        \end{minipage}%
        \hspace{\margin}%
        \begin{minipage}[t]{\txtw}
            \vspace{0pt} % Essential for top alignment with images
            \raggedright
            The image depicts an egg that has been creatively styled to resemble the famous painting "Girl with a Pearl Earring" by Johannes Vermeer. The egg is adorned with a blue headscarf and a yellow drape, mimicking the attire of the girl in the painting. Additionally, a pearl-like object is attached to the egg, representing the iconic pearl earring from the original artwork. The background is a rich blue, similar to the background in Vermeer's painting, further enhancing the reference. This playful and artistic representation pays homage to the classic piece of art in a whimsical and imaginative manner.
        \end{minipage}%
        \hfill %
    \end{subfigure}

    % --- ROW 2 ---
    \vspace{\rowsep}
    \begin{subfigure}[t]{\linewidth}
        \begin{minipage}[t]{\capw}
            \caption{}
            \label{fig:rethink_plastic}
        \end{minipage}%
        \begin{minipage}[t]{\imgw}
            \vspace{0pt} % Essential for top alignment with images
            \includegraphics[width=\linewidth]{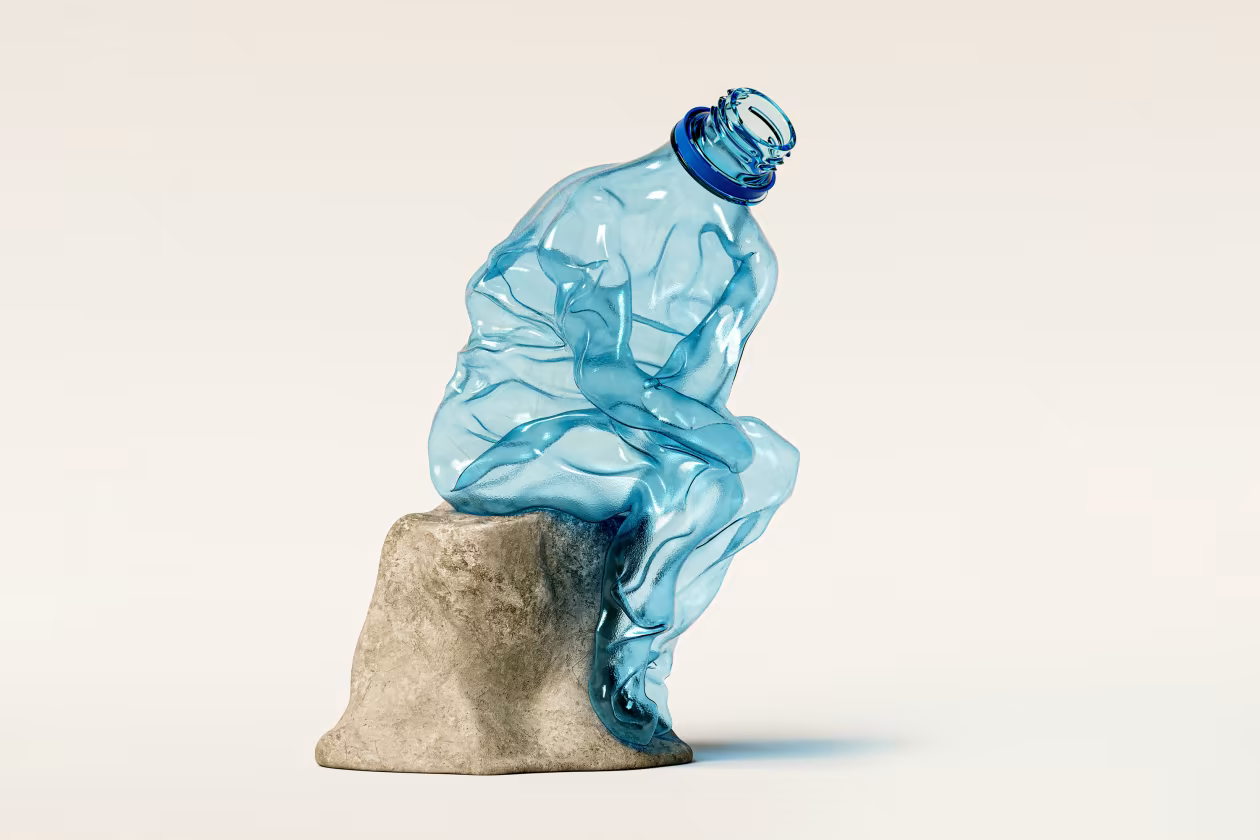}
            {\centering \hyperlink{https://www.barrons.com/articles/cheap-new-plastic-choking-the-world-9b318936}{``Rethink Plastic''}\cite{jaen2023rethink} \\ \copyright \hyperlink{https://javierjaen.com/20-2-23-Barron-s-Rethinking-Plastic}{Javier Jaén} \par
            \weeny{Used with permission.} \par}
        \end{minipage}%
        \hspace{\margin}%
        \begin{minipage}[t]{\txtw}
            \vspace{0pt} % Essential for top alignment with images
            \raggedright
            This image is a creative and thought-provoking piece of art. It depicts a crumpled blue plastic bottle that has been shaped to resemble ``The Thinker,'' a famous sculpture by Auguste Rodin. The original sculpture portrays a man in deep contemplation, often used to symbolize philosophy and introspection. \par
            
            In this artwork, the plastic bottle mimics the pose of ``The Thinker,'' sitting on a stone-like base with its head resting on one hand and the other arm resting on its knee. The use of a plastic bottle, a common symbol of environmental pollution, adds a layer of commentary on modern issues such as waste, recycling, and the impact of human activity on the environment. The juxtaposition of a classical art form with a contemporary environmental issue creates a powerful visual statement.
        \end{minipage}%
        \hfill %
    \end{subfigure}

    % --- ROW 3 ---
    \vspace{\rowsep}
    \begin{subfigure}[t]{\linewidth}
        \begin{minipage}[t]{\capw}
            \caption{}
            \label{fig:starry_night_on_a_pcb}
        \end{minipage}%
        \begin{minipage}[t]{\imgw}
            \vspace{0pt} % Essential for top alignment with images
            \includegraphics[width=\linewidth]{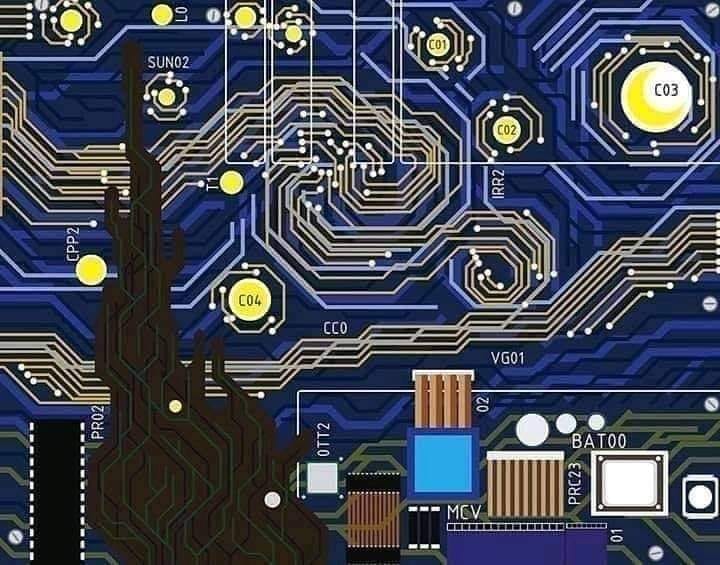}
            {\centering \hyperlink{https://www.behance.net/gallery/41188523/Van-Gogh-on-a-PCB}{``Van Gogh on a PCB''}\cite{takasu2016vangogh} \\ \copyright \hyperlink{https://www.behance.net/tkasu}{Tarik Takasu} \\ \weeny{\textit{Via \hyperlink{https://www.behance.net}{Behance}. \hyperlink{https://www.behance.net/gallery/41188523/Van-Gogh-on-a-PCB}{Licensed under CC BY-NC-ND 4.0.}}} \par}
        \end{minipage}%
        \hspace{\margin}%
        \begin{minipage}[t]{\txtw}
            \vspace{0pt} % Essential for top alignment with images
            \raggedright
            This image is a creative and artistic representation that combines elements of a printed circuit board (PCB) with the famous painting ``The Starry Night'' by Vincent van Gogh. \par
            Key features include:
            \begin{itemize}[leftmargin=*]
                \setlength{\itemsep}{0pt}
                \setlength{\parskip}{0pt}
                \setlength{\parsep}{0pt}
                \item \textit{Circuit Board Elements}: The image incorporates typical components and traces found on a PCB, such as labeled nodes (e.g., C01, C02, C03), pathways, and connectors. The labels and the structured layout are characteristic of electronic circuit designs.
                \item \textit{Artistic Elements}: The swirling patterns and the arrangement of the yellow circles resemble the stars and the swirling night sky in van Gogh's ``The Starry Night.'' The dark, tree-like structure on the left side of the image mimics the cypress tree in the original painting.
                \item \textit{Color Scheme}: The predominant use of blue and yellow colors aligns with the color palette of ``The Starry Night,'' enhancing the visual connection between the two themes.
            \end{itemize}
            This fusion of technology and classic art creates a unique and visually striking piece that pays homage to both the intricacies of electronic design and the timeless beauty of van Gogh's masterpiece.
        \end{minipage}%
        \hfill %
    \end{subfigure}

    % --- ROW 4 ---
    \vspace{\rowsep}
    \begin{subfigure}[t]{\linewidth}
        \begin{minipage}[t]{\capw}
            \caption{}
            \label{fig:sink_draining}
        \end{minipage}%
        \begin{minipage}[t]{\imgw}
            \vspace{0pt} % Essential for top alignment with images
            \includegraphics[width=\linewidth]{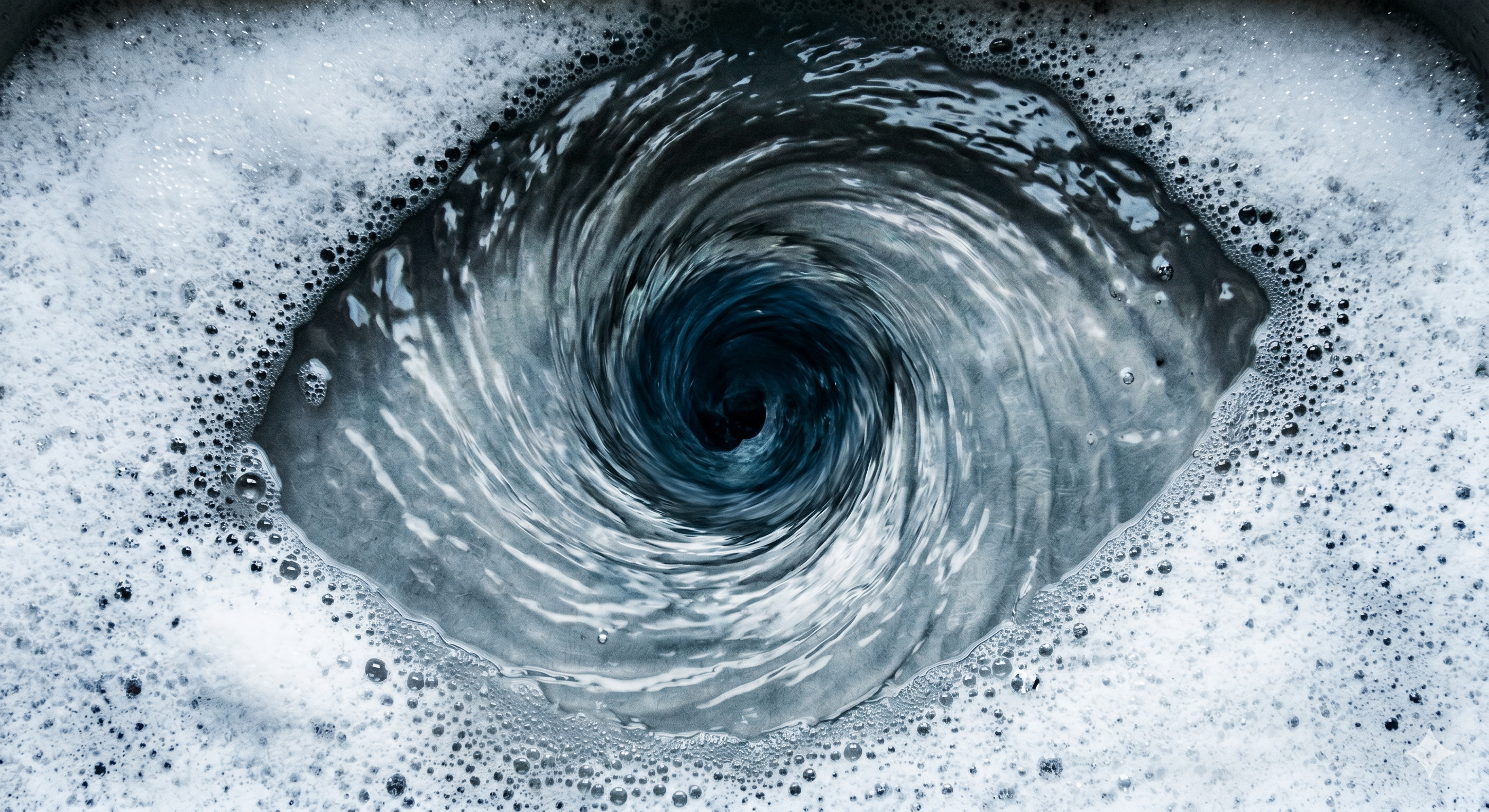}
            {\centering Generated by Google Gemini \par}
        \end{minipage}%
        \hspace{\margin}%
        \begin{minipage}[t]{\txtw}
            \vspace{0pt} % Essential for top alignment with images
            \raggedright
            This image depicts a whirlpool in a body of soapy water, such as a sink or a bathtub. The swirling motion of the water creates a vortex, drawing the viewer's eye to the center. The bubbles and foam around the edges of the whirlpool add texture and context, indicating that the water is soapy.
            The image is visually striking because the whirlpool resembles an eye, with the swirling water forming the iris and the darker center of the vortex resembling a pupil. This visual metaphor can evoke a sense of being watched or drawn into the image, creating a feeling of depth and intrigue.
            The reference to an eye might also suggest themes of perception, observation, or introspection. The swirling motion can symbolize the idea of being pulled into a situation or the cyclical nature of certain experiences.
            Overall, the image combines elements of nature and everyday life to create a visually compelling and thought-provoking scene.
        \end{minipage}%
        \hfill %
    \end{subfigure}
    \vspace{-6pt}
    \caption{Qualitative assessment of the ability of GPT-4o in understanding art content on challenging handpicked examples. The prompt used was ``Tell me about this image fully. Identify any references it is making.''.}
    \squeezegap
    \label{fig:gpt4o-qualitative-best}
\end{figure}

Despite the officially admitted limitations, we found GPT-4o to be proficient in \textit{artistic comprehension}. Given how art often expresses concepts through abstractions, motifs and various creative means, art-comprehension can be particularly challenging, even to humans. We are not aware of prior work that specifically assesses MLLMs on art comprehension. To give a sense of GPT-4o's capability, we present its responses on some challenging curated images in~\cref{fig:gpt4o-qualitative-best}. Notably, \cref{fig:girl_with_a_pearl_earring,fig:rethink_plastic,fig:starry_night_on_a_pcb} reference, respectively, Vermeer's \emph{Girl with a Pearl Earring}, Rodin's sculpture \emph{The Thinker}, and van Gogh's \emph{The Starry Night}." In addition to assessing whether the MLLM can ``perceive'' past the visual abstractions, these images are also a test of the MLLM's knowledge of art since the references in question are of famous artworks. GPT-4o not only correctly identified the references and explained their salient motifs but also inferred the social commentary that \cref{fig:rethink_plastic} makes on plastic waste, despite not being given the artwork's actual title, ``Rethink Plastic.'' Impressively, when provided with an optical illusion in~\cref{fig:sink_draining}, the MLLM accurately identified the swirling whirlpool and how it creates the visual impression of an eye. We also observe similar capability in the CompArt annotations. For example, the annotation example for ``Caption'' in \cref{fig:compart-gallery-caption} shows Francis Picabia's \emph{Dances at the Spring}, a cubist depiction of two dancing figures; GPT-4o correctly perceives it as ``two abstract human figures.''

\begin{figure}[t]
    \captionsetup[subfigure]{labelformat=empty}
    \scriptsize
    \centering
    % Row 1
    \begin{subfigure}[b]{0.195\linewidth}
        \centering
        \includegraphics[width=\linewidth]{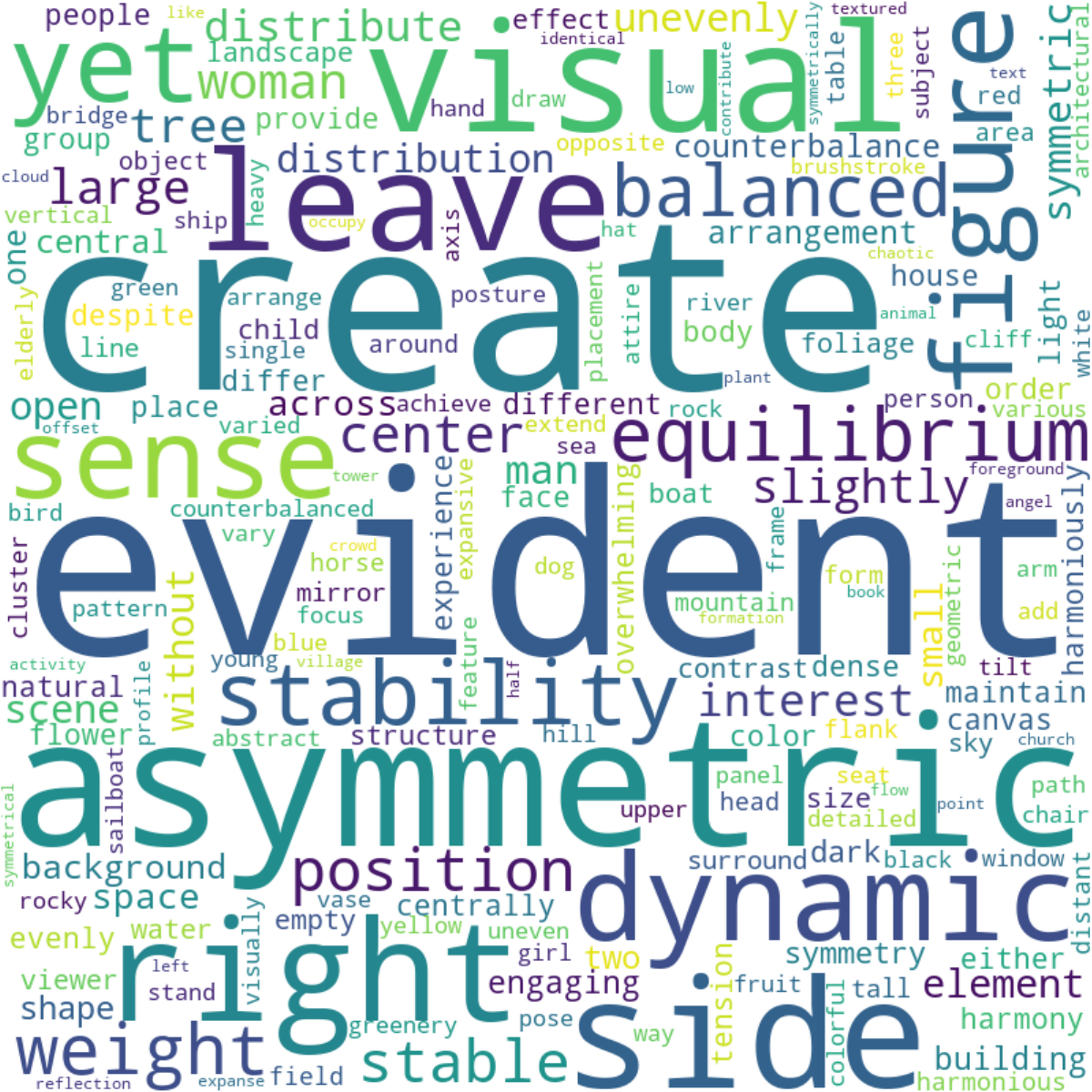}
        \caption{Balance}
    \end{subfigure}
    \begin{subfigure}[b]{0.195\linewidth}
        \centering
        \includegraphics[width=\linewidth]{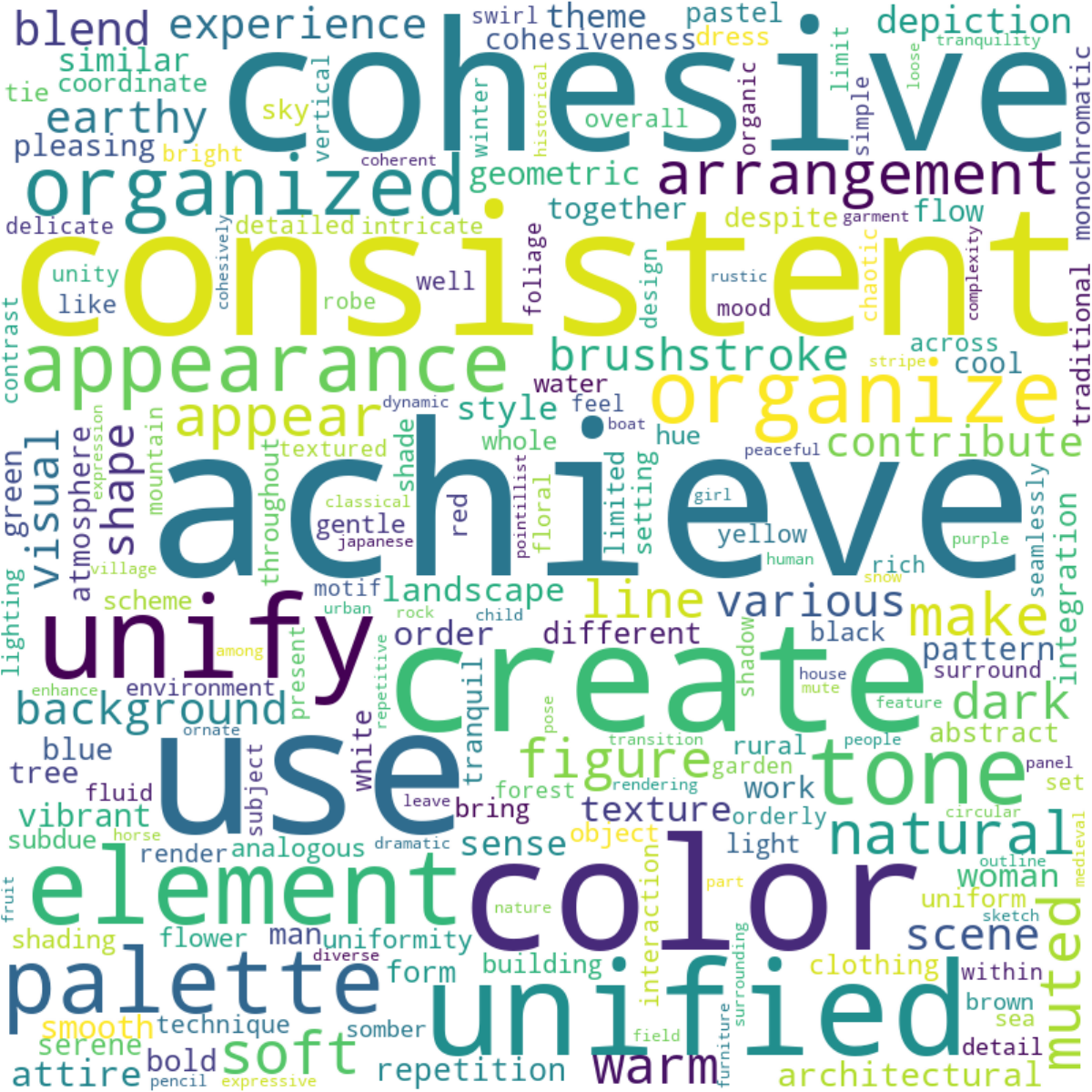}
        \caption{Harmony}
    \end{subfigure}
    \begin{subfigure}[b]{0.195\linewidth}
        \centering
        \includegraphics[width=\linewidth]{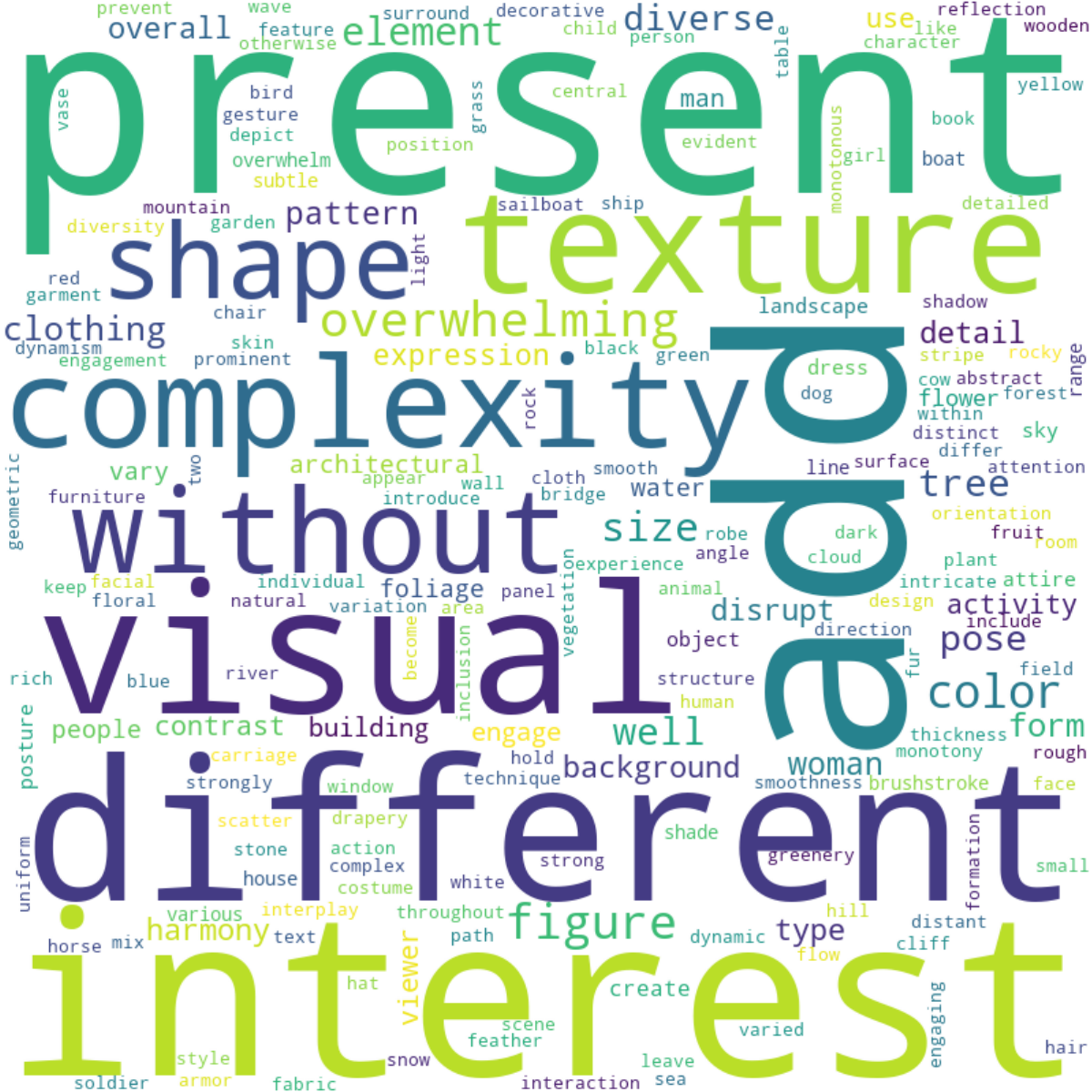}
        \caption{Variety}
    \end{subfigure}
    \begin{subfigure}[b]{0.195\linewidth}
        \centering
        \includegraphics[width=\linewidth]{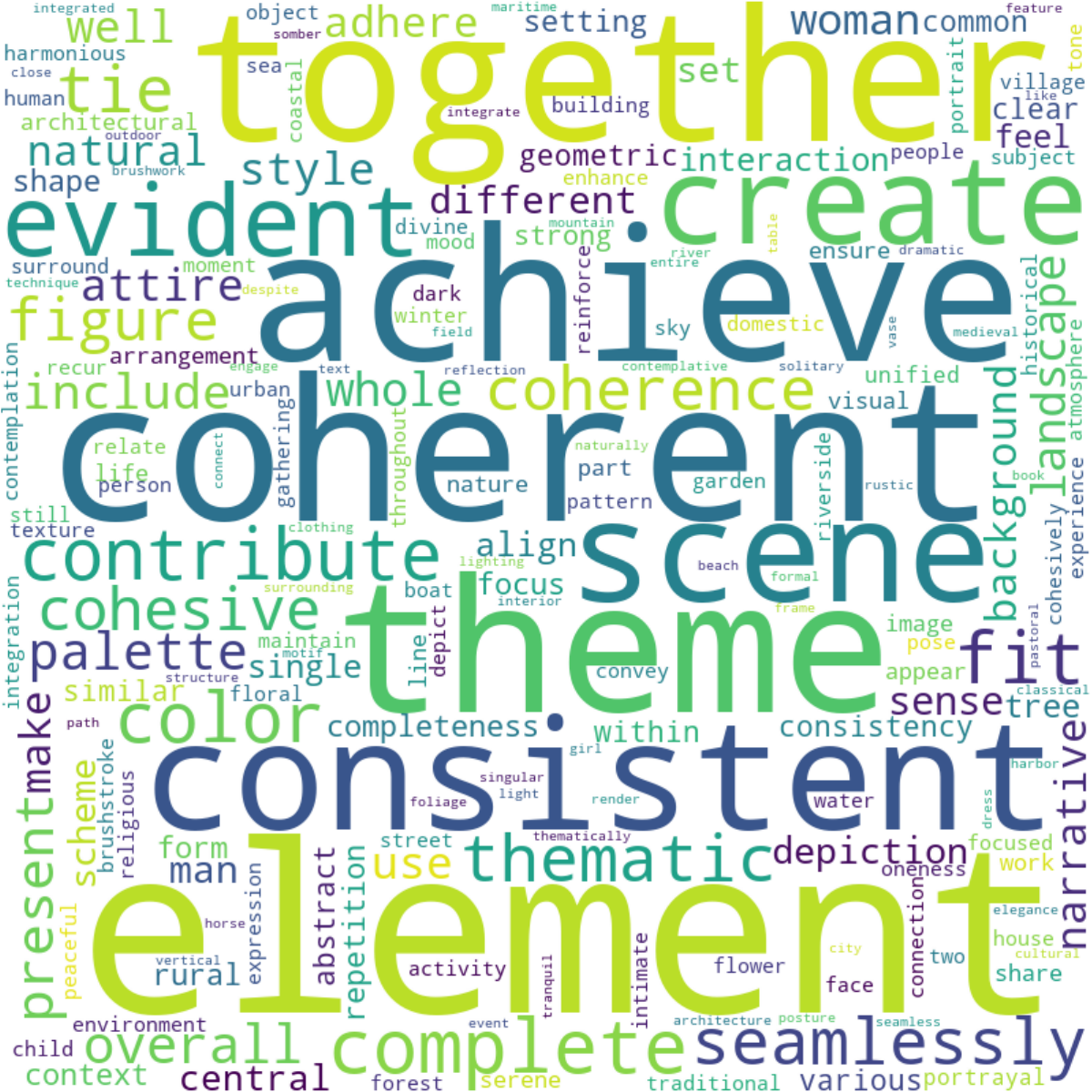}
        \caption{Unity}
    \end{subfigure}
    \begin{subfigure}[b]{0.195\linewidth}
        \centering
        \includegraphics[width=\linewidth]{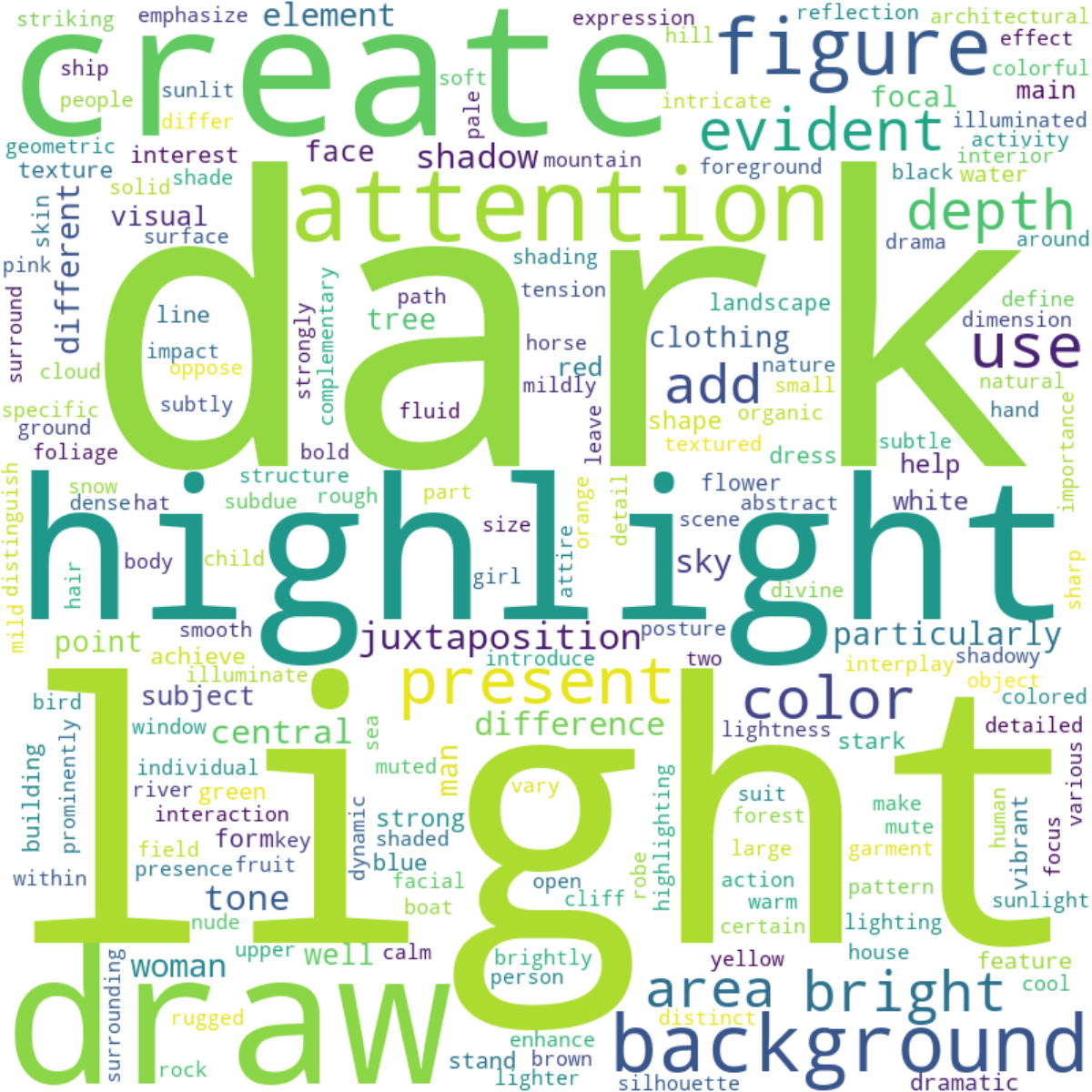}
        \caption{Contrast}
    \end{subfigure}
    % Row 2
    \begin{subfigure}[b]{0.195\linewidth}
        \centering
        \includegraphics[width=\linewidth]{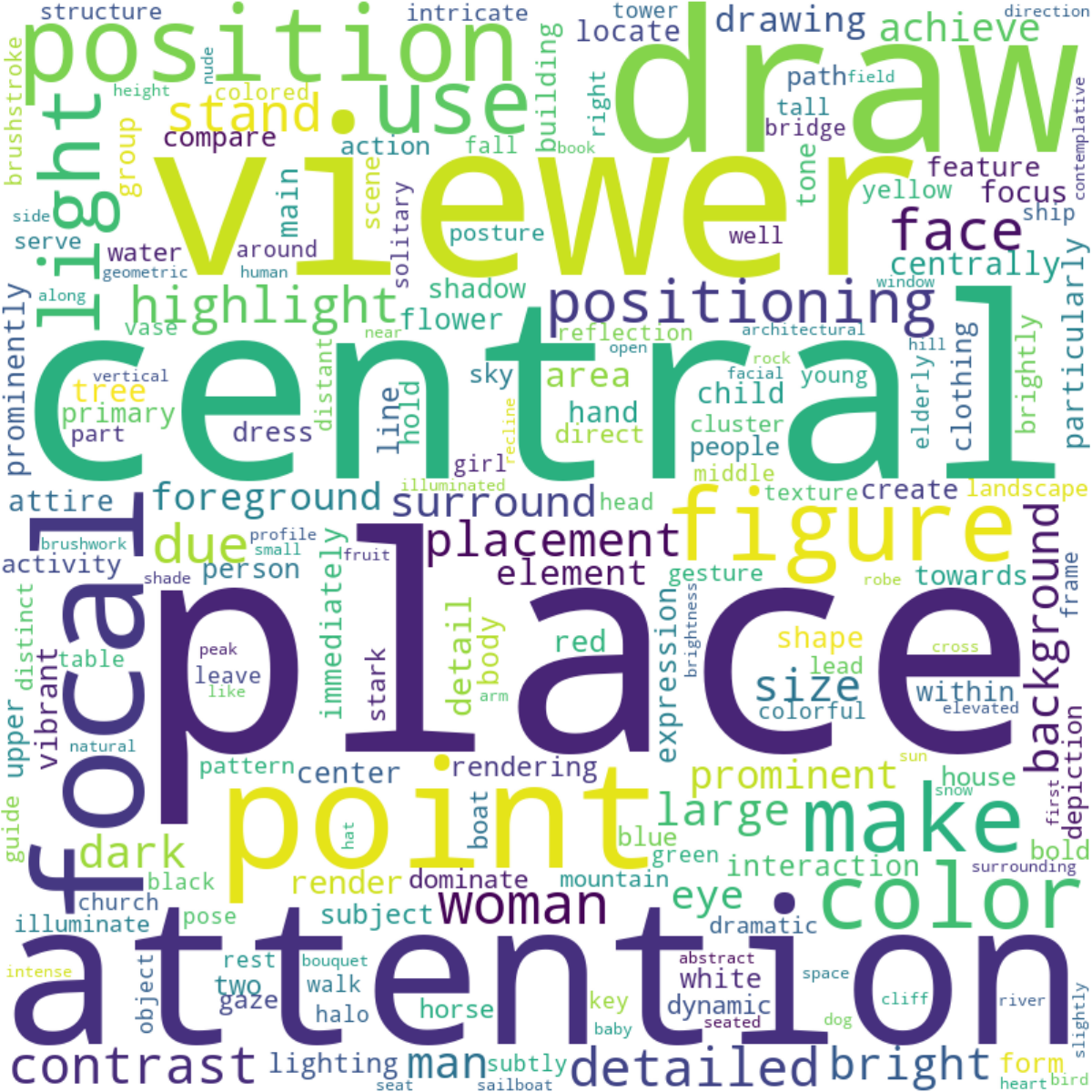}
        \caption{Emphasis}
    \end{subfigure}
    \begin{subfigure}[b]{0.195\linewidth}
        \centering
        \includegraphics[width=\linewidth]{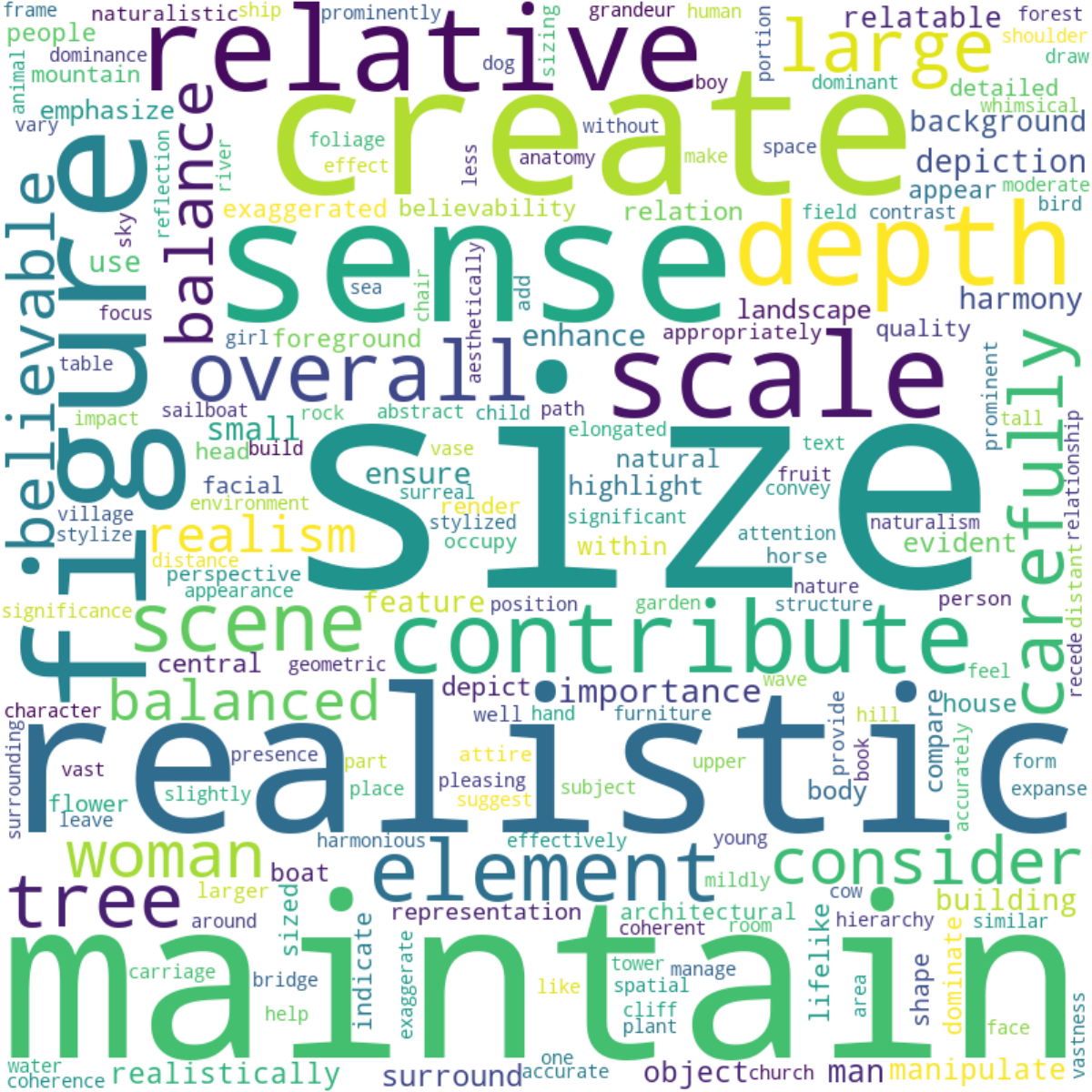}
        \caption{Proportion}
    \end{subfigure}
    \begin{subfigure}[b]{0.195\linewidth}
        \centering
        \includegraphics[width=\linewidth]{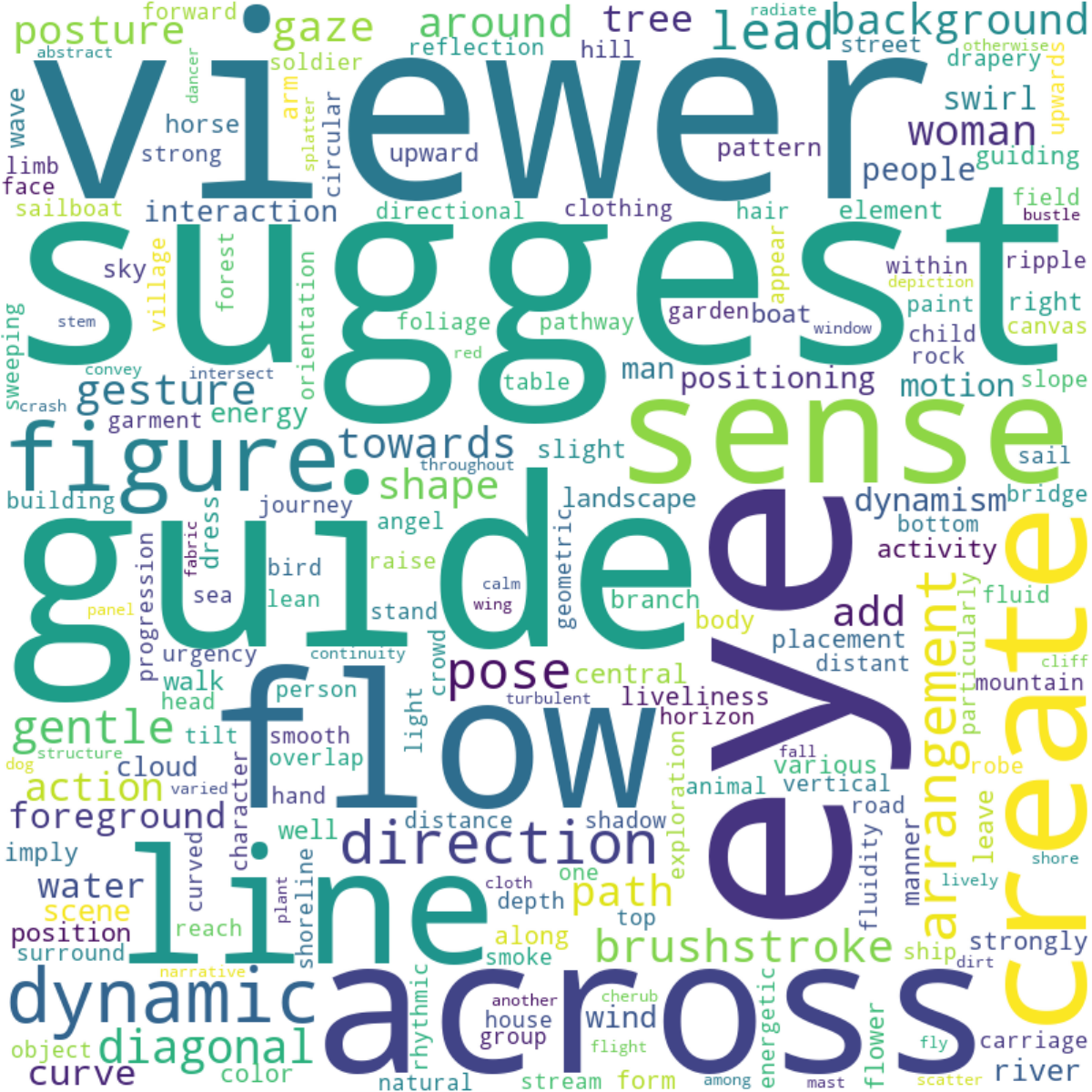}
        \caption{Movement}
    \end{subfigure}
    \begin{subfigure}[b]{0.195\linewidth}
        \centering
        \includegraphics[width=\linewidth]{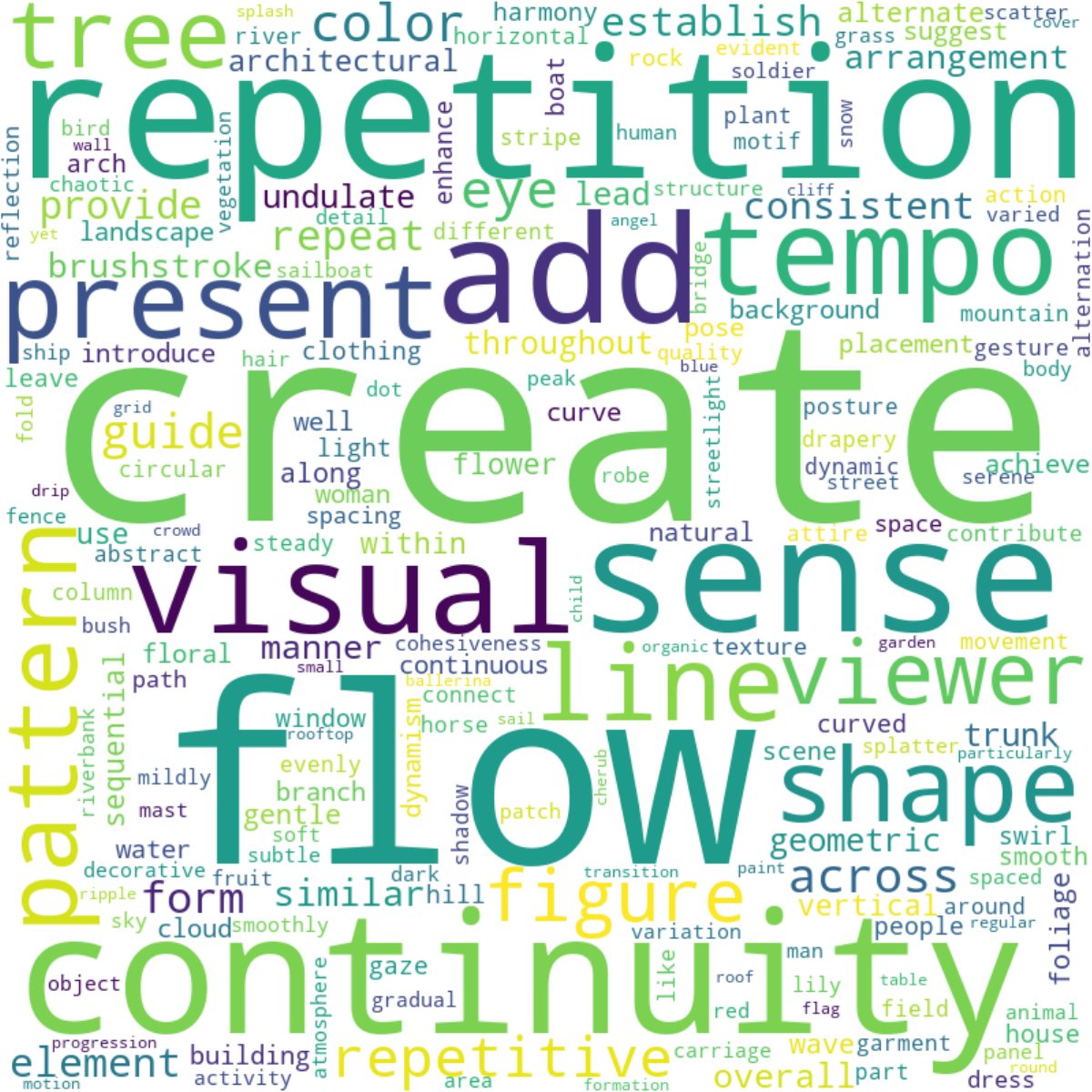}
        \caption{Rhythm}
    \end{subfigure}
    \begin{subfigure}[b]{0.195\linewidth}
        \centering
        \includegraphics[width=\linewidth]{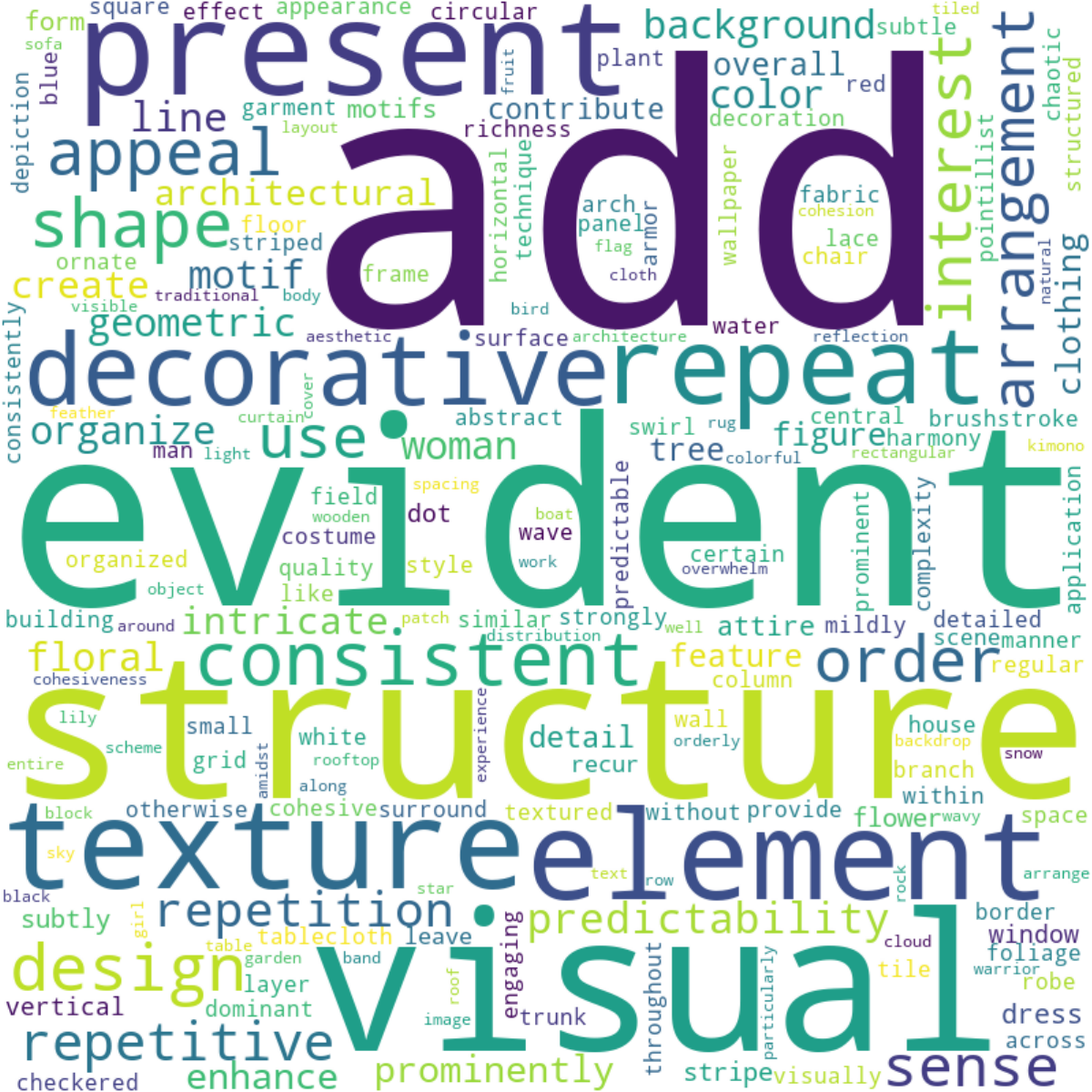}
        \caption{Pattern}
    \end{subfigure}
    \caption{Word clouds of each PoA annotation type based on observed term frequencies after lemmatization. Stopwords, names of PoA principles and ``composition'' term are ignored.}
    \squeezegap
    \label{fig:word-clouds}
\end{figure}

\textbf{Annotation quality}. We also qualitatively assess the \textit{linguistic correctness} of PoA annotations in CompArt. We first examine whether principles differ in content. As shown in the lower table of~\cref{tab:richness_diversity}, varying PoS score distributions reflect each principle's unique linguistic makeup. For example, the Movement principle has higher verb scores than the Pattern principle, which describes more static compositional elements. Next, Word clouds of top annotation terms per principle (\cref{fig:word-clouds}) reveal subtle conceptual distinctions. For example, Harmony and Unity principles overlap on ``theme'' but diverge in sense: Harmony tends toward ``cohesive'', while Unity tends toward ``coherent''. Similarly, while both Rhythm and Pattern principles are about ``repetition'', the former describes a more dynamic ``flow'' while the latter is about describing static ``structure''. These strongly suggest that the annotations do not suffer from hallucinations at the language level and they exhibit strict adherence to our PoA definitions.

% Dataset/method (incl analysis): 3-3.5 pages. Leading paragraph to introduce subsections/studies.
% for method and results, if subsections exist, there must be a leading paragraph to describe them
\section{Method}
In this section, we first present the preliminaries to T2I generation (\cref{sec:method_prelim}) followed by our design of ArtDapter (\cref{sec:method_artdapter}), detailing the input featurization, condition injection flow and its architectural design.

\subsection{Preliminary}
\label{sec:method_prelim}
We build on Stable Diffusion (SD)~\cite{Rombach2021HighResolutionIS}, a latent diffusion model~\cite{DBLP:conf/nips/HoJA20} that denoises a Gaussian prior toward the target data distribution. SD performs both the forward and backward processes in a latent space using a U-Net backbone, and injects textual conditions into the backward process via cross-attention, $\text{Attention}(Q,K,V) = \text{Softmax}(\frac{QK^T}{\sqrt{d}}) \cdot V$, such that
\begin{equation}
  \label{eq:crossattn}
  Q=W_q(z),\quad K=W_k(y),\quad V=W_v(y),
\end{equation}
where $z$ is the noisy latent, $y$ is the text token embeddings from a text encoder (e.g., CLIP), and $W_q, W_k, W_v$ are projection matrices. ArtDapter intervenes at this text-conditioning interface (Section~\ref{sec:method_artdapter}).

\subsection{ArtDapter}
\label{sec:method_artdapter}
\begin{figure}[H]
  \centering
  \includegraphics[width=.8\linewidth]{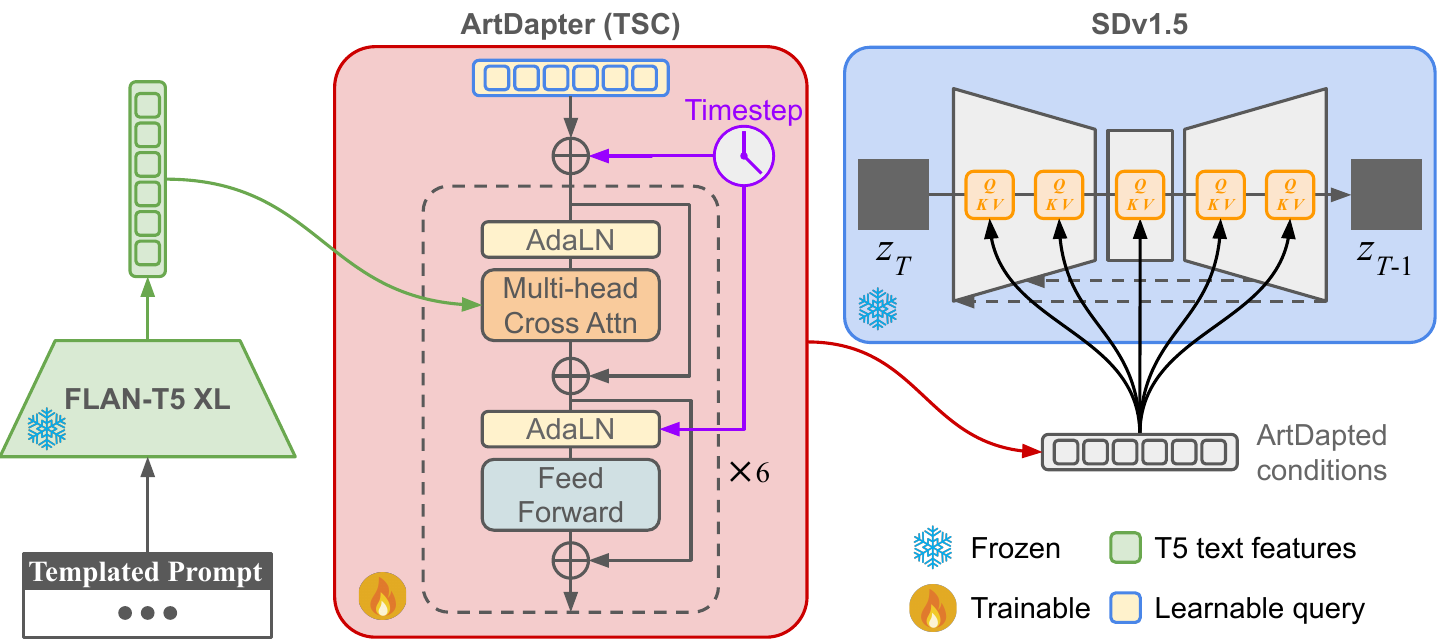}
  \caption{Model overview. The templated prompt is first transformed into text features using FLAN-T5 XL before projected to ArtDapted conditions via ArtDapter. Cross-attention is used to inject the ArtDapted conditions into the DM. Crucially, only the TSC module is trainable.}
  \squeezegap
  \label{fig:artdapter}
\end{figure}

\textbf{ArtDapter} is our adapter for latent DMs that conditions T2I generation on user-specified PoA controls. It is an adapter for latent DMs. Below we describe how PoA controls are encoded and injected into the diffusion process; \cref{fig:artdapter} illustrates the framework.

\textbf{Capturing PoA conditions.} PoA annotations are inherently \textit{semi-global}: they describe an artwork's spatial contents and their effect on global composition, but without explicit localization. (examples in~\cref{fig:compart-gallery}). We leverage the rich representational capacity of LLMs to encode the long, semantically dense PoA annotations. Given a condition tuple of (caption, art-style, PoA controls) at train or inference time, we concatenate the components in a fixed order to form a single \emph{templated prompt}. Missing components are filled with the literal token \texttt{"None"}. We then encode this prompt using the LLM to obtain contextual text features. Packing all components into one sequence encourages the encoder to model their relationships jointly (e.g., how a PoA constraint should manifest under a specific style and caption).

\textbf{Projection architecture.} ArtDapter takes the LLM features as input and projects them into the prompt-token space of SD, where they are injected via cross-attention. This design lets ArtDapter transfer across DMs that share a similar U-Net backbone. For its architecture, we adopt the Timestep-Aware Semantic Connector from ELLA~\cite{Hu2024ELLAED}: a six-block resampler with timestep integration in the Adaptive Layer Normalization~\cite{DBLP:conf/iccv/PeeblesX23}. In total, ArtDapter has 66.82M learnable parameters — roughly 18\% of ControlNet's 361M~\cite{DBLP:conf/iccv/ZhangRA23} — while exposing 10 compositional controls (vs.\ ControlNet's structural ones).

% Evaluation & Results: 2 pages
% for method and results, if subsections exist, there must be a leading paragprah to describe them

% \newcolumntype{P}[1]{>{\centering\arraybackslash}p{#1}}
\begin{figure}[t]
    \centering
    \includegraphics[width=\linewidth]{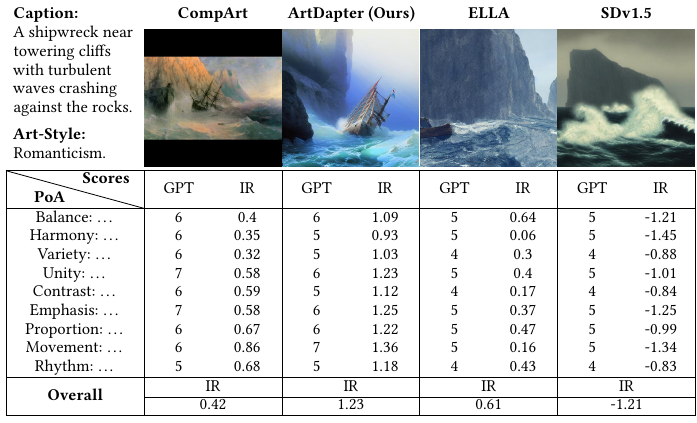}
    \caption{Our evaluation scorecard from an actual test sample. The left column details the artistic conditions (PoA conditions truncated) for the associated CompArt image and the generations of ArtDapter, ELLA and SDv1.5 outputs. For each image, we score its alignment to \textit{each} PoA condition using GPT-4o (GPT) and ImageReward~\cite{DBLP:conf/nips/XuLWTLDTD23} (IR).}
    \squeezegap
    \label{fig:scorecard-example}
\end{figure}

% GPT is instructed to score on the 7-point Likert Scale~\cite{likert1932measurement} with 1 being poor alignment and 7 being excellent alignment. The GPT prompt used is included in the Supplementary Materials~\cref{sec:evaluation_prompt}. The IR scoring prompt is obtained by concatenating the caption with the PoA condition in question. To decide the winner of each PoA condition alignment, we simply pick the image with the highest GPT score. If there is a tie, the one with higher IR score is chosen. The overall winner for the generation is the image with the most PoA alignment wins. For the given example, if CompArt was included in the evaluation round, then it is the overall winner with 5 PoA alignment wins. If we did not include CompArt, then the overall winner is ArtDapter with 9 PoA alignment wins.

\section{Experiments}
This section details our training scheme, baselines, and evaluation framework.

\subsection{Training details}
\label{sec:training_details}

\textbf{Disentanglement Strategy.} Because PoA principles are not mutually exclusive concepts, we employ a joint training scheme that enables the model to learn both single-PoA generation and the effective composition of multiple PoA conditions. This is realized through each training sample by employing a 50\% probability to drop the caption, a 50\% probability to drop each PoA, a 10\% probability to drop all PoA and a 10\% probability to keep all PoA. To ensure ArtDapter learns generalized compositional rules rather than memorizing style-specific correlations, the \textit{art-style} condition is never dropped. This forces the model to learn PoA concepts as modulators \textit{on top of} various styles, effectively disentangling composition from style and semantics.

\textbf{Configurations.} We use the Flan T5-XL~\cite{DBLP:journals/jmlr/RaffelSRLNMZLL20} text encoder to obtain ArtDapter input from the templated prompt. While any capable LLM would be appropriate, T5-XL helps us establish a fair comparison against ELLA~\cite{Hu2024ELLAED}. In ensuring the comparability of our results and the transferability of ArtDapter to open-sourced DMs, we adopt the pre-trained SDv1.5~\cite{Rombach2021HighResolutionIS} for our T2I DM. We use CompArt's train-split for training, resizing the inputs to $512 \times 512$. Training spans 245,000 iterations, optimized using AdamW~\cite{DBLP:journals/corr/KingmaB14} with a weight decay of 0.01 and a learning rate of $1 \times 10^{-5}$. During inference, we employ DDIM~\cite{DBLP:conf/iclr/SongME21} as our sampling strategy with 50 steps and CFG~\cite{ho2021classifierfree} scale set to 7.5. This sampling configuration is maintained across all models during evaluation.

\subsection{Evaluation setup}
\label{sec:evaluation_setup}

\input{fig/evaluation_results}

Using the CompArt test split (1,000 sets of art conditions), we generate outputs with our trained \textit{ArtDapter}, \textit{ELLA}~\cite{Hu2024ELLAED} model pre-trained for semantic-alignment, and the pre-trained \textit{SDv1.5} model. The latter two provide comparable baselines, and we used their publicly available weights. Since neither ELLA nor SDv1.5 is trained to accept templated prompts, we instead concatenate the caption, art-style, and all PoAs (in fixed sequence) into a single long prompt. Thus, their outputs reflect only the models' inherent understanding of the given art-styles and PoA conditions in plain text.

\subsection{Evaluation scheme}
\label{sec:evaluation_scheme}

\textbf{Protocol (overview).}
For each test condition set, we: (1) generate one image per model, (2) score each image for each requested PoA principle via a MLLM rubric, (3) aggregate principle-level wins, and (4) use ImageReward only to break ties and to provide an external signal.

Evaluating compositional adherence is challenging due to the lack of ground-truth metrics for subjective qualities. We acknowledge the potential circularity of using the annotator (GPT-4o) as the evaluator. To address this, we design a dual-metric framework:

\begin{itemize}[leftmargin=*]
    \item \textbf{Consistency Check (MLLM-based):} With the same PoA definitions for annotating CompArt, we employed GPT-4o to score how well a generated image matches its caption/style/PoA controls. This measures alignment with the \emph{operationalized PoA specification} rather than claiming human-ground-truths.
    \item \textbf{External Signal (ImageReward):} To mitigate annotator bias, we employ ImageReward (IR)~\cite{DBLP:conf/nips/XuLWTLDTD23}, a preference model trained on human feedback, as an external arbiter. IR serves as a tie-breaker and independent quality check.
\end{itemize}

\begin{figure}[t]
    % Post-Impressionism', 'Expressionism', 'Impressionism', 'Northern Renaissance', 'Realism', 'Romanticism', 'Symbolism', 'Art Nouveau (Modern)', 'Naïve Art (Primitivism)', 'Baroque', 'Rococo', 'Abstract Expressionism', 'Cubism', 'Color Field Painting', 'Pop Art', 'Pointillism', 'Early Renaissance', 'Ukiyo-e', 'Mannerism (Late Renaissance)', 'High Renaissance', 'Fauvism', 'Minimalism', 'Action painting', 'Contemporary Realism', 'Synthetic Cubism', 'New Realism', 'Analytical Cubism'
    \centering
    \weeny
    \begin{subfigure}[]{\linewidth}
    \setlength{\tabcolsep}{0pt}
        \begin{tabular}{c c c c c c c c c}
             \includegraphics[width=.111\textwidth,height=.111\textwidth]{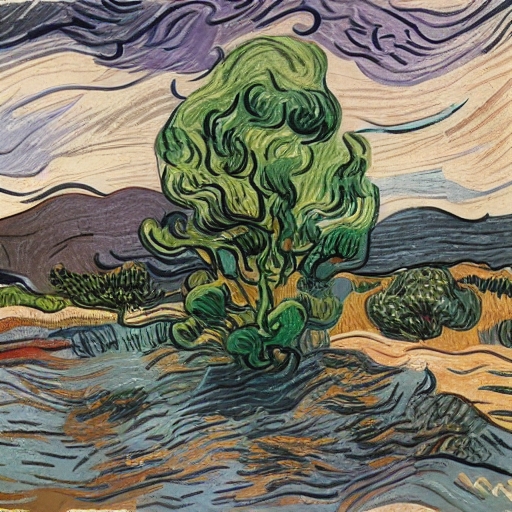} & % Post-Impressionism
             \includegraphics[width=.111\textwidth,height=.111\textwidth]{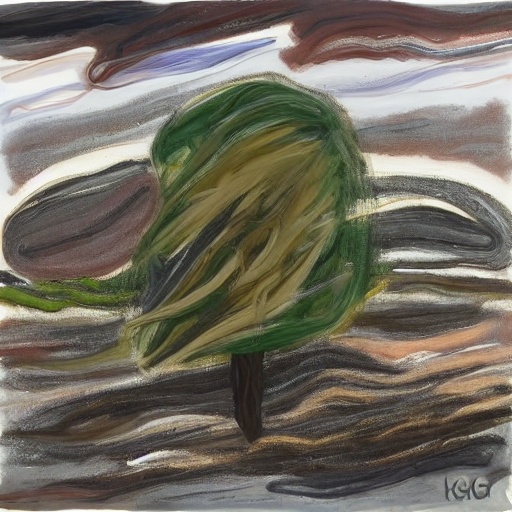} & % Expressionism
             \includegraphics[width=.111\textwidth,height=.111\textwidth]{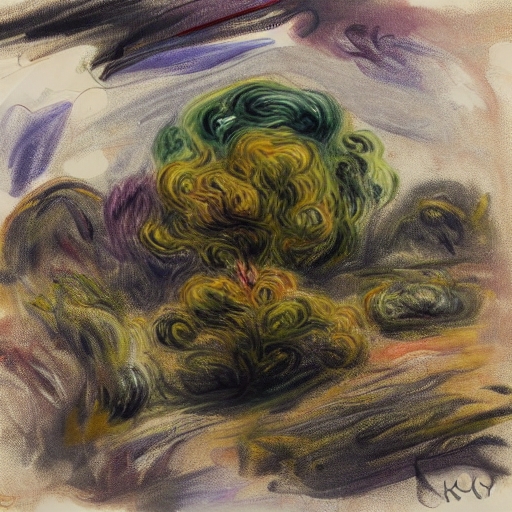} & % Impressionism
             \includegraphics[width=.111\textwidth,height=.111\textwidth]{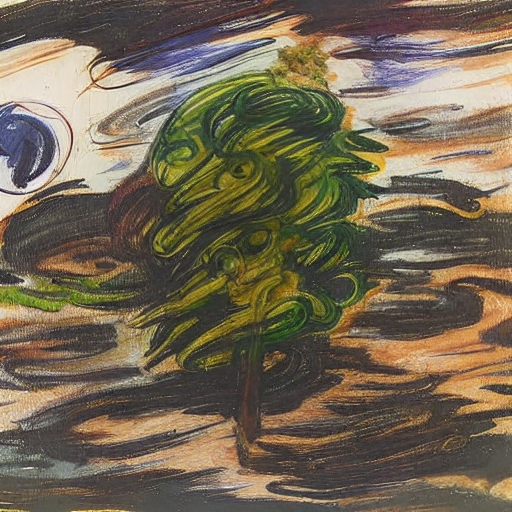} & % Northern Renaissance
             \includegraphics[width=.111\textwidth,height=.111\textwidth]{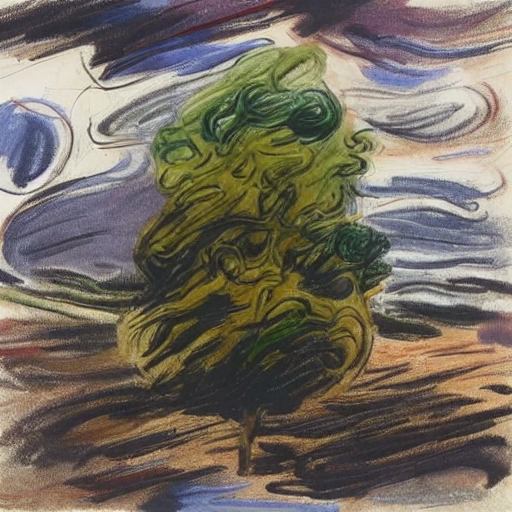} & % Realism
             \includegraphics[width=.111\textwidth,height=.111\textwidth]{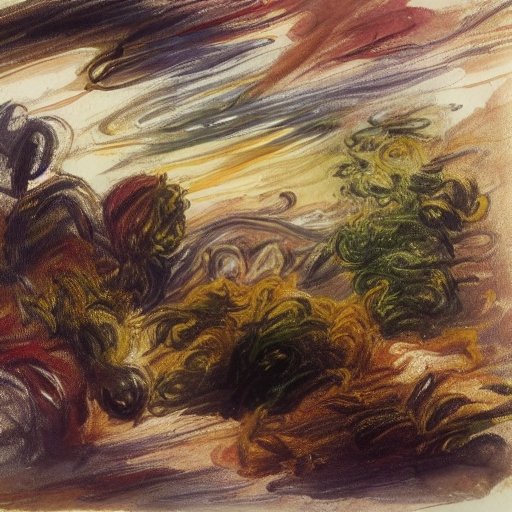} & % Romanticism
             \includegraphics[width=.111\textwidth,height=.111\textwidth]{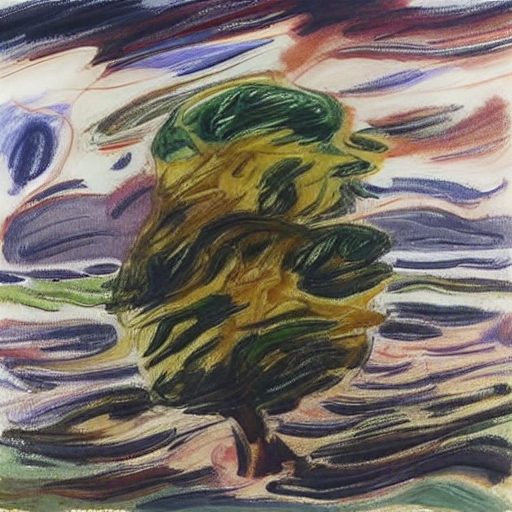} & % Symbolism
             \includegraphics[width=.111\textwidth,height=.111\textwidth]{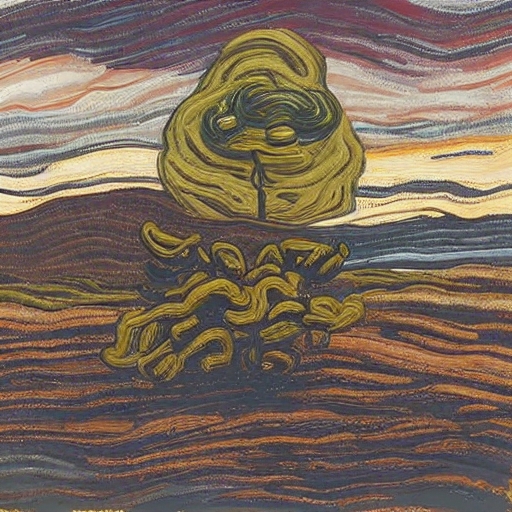} & % Art Nouveau (Modern)
             \includegraphics[width=.111\textwidth,height=.111\textwidth]{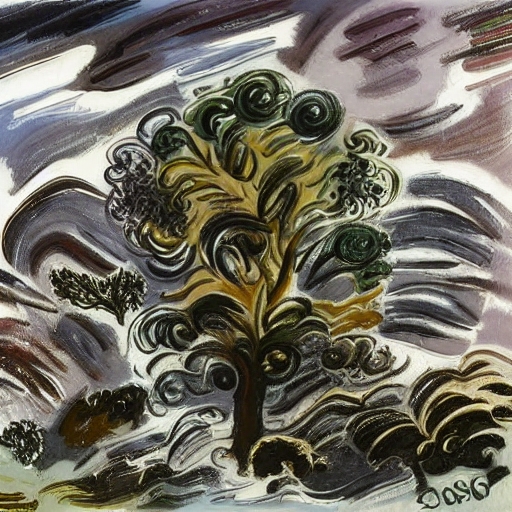} \\ % Naïve Art (Primitivism)
            \makecell{Post-\\Impressionism} &
            {Expressionism} &
            {Impressionism} &
            \makecell{Northern\\Renaissance} &
            {Realism} &
            {Romanticism} &
            {Symbolism} &
            \makecell{Art Nouveau\\(Modern)} &
            \makecell{Naïve Art\\(Primitivism)}\\
            \includegraphics[width=.111\textwidth,height=.111\textwidth]{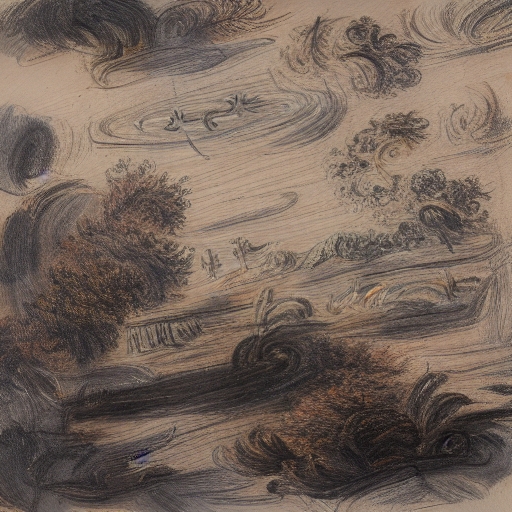} & % Baroque
            \includegraphics[width=.111\textwidth,height=.111\textwidth]{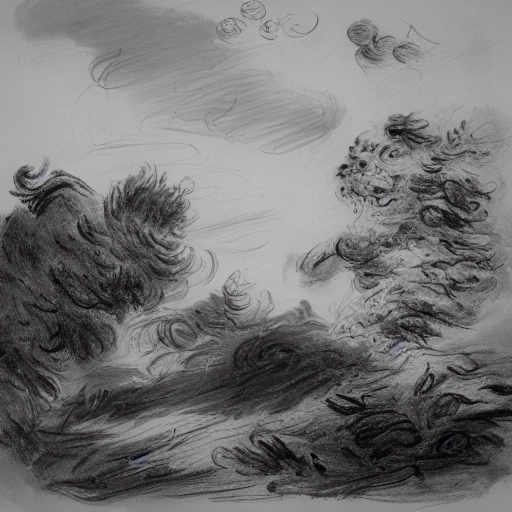} & % Rococo
            \includegraphics[width=.111\textwidth,height=.111\textwidth]{assets/style_wise/vincent-van-gogh_trees-and-shrubs-1889-1_abstract-expressionism.jpg} & % Abstract Expressionism
            \includegraphics[width=.111\textwidth,height=.111\textwidth]{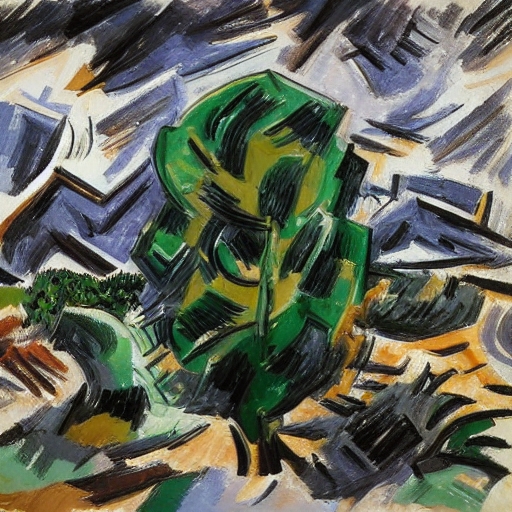} & % Cubism
            \includegraphics[width=.111\textwidth,height=.111\textwidth]{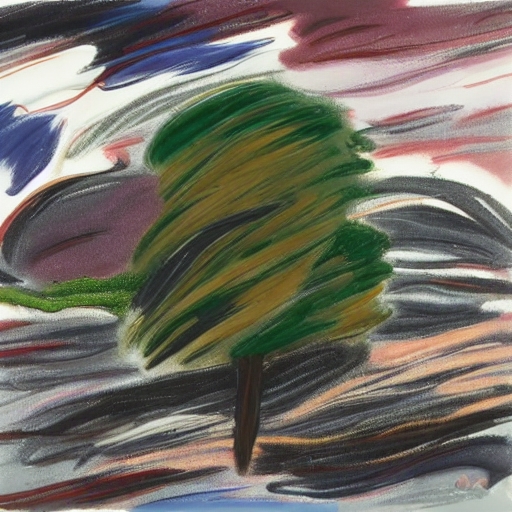} & % Color Field Painting
            \includegraphics[width=.111\textwidth,height=.111\textwidth]{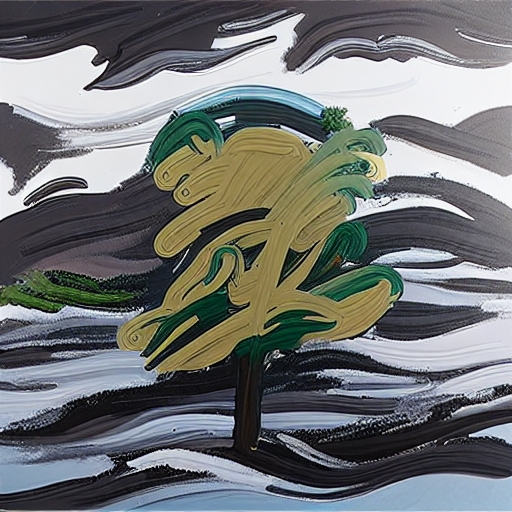} & % Pop Art
            \includegraphics[width=.111\textwidth,height=.111\textwidth]{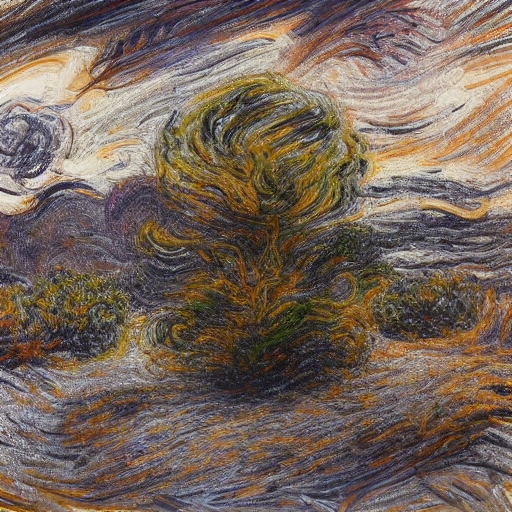} & % Pointillism
            \includegraphics[width=.111\textwidth,height=.111\textwidth]{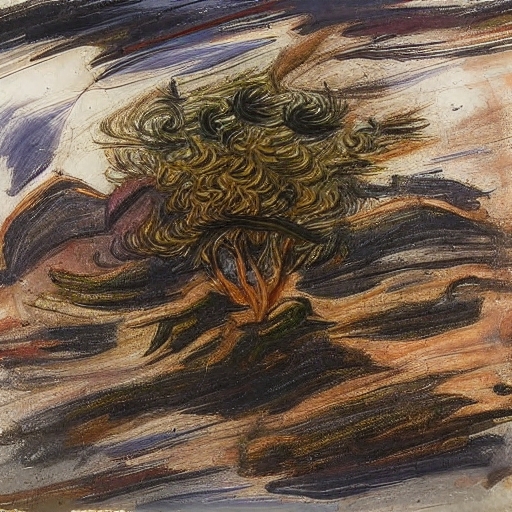} & % Early Renaissance
            \includegraphics[width=.111\textwidth,height=.111\textwidth]{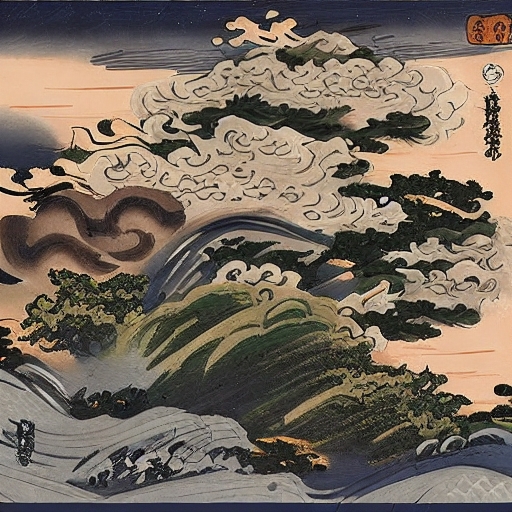}\\ % Ukiyo-e
            {Baroque} &
            {Rococo} &
            \makecell{Abstract\\Expressionism} &
            {Cubism} &
            \makecell{Color Field\\Painting} &
            {Pop Art} &
            {Pointillism} &
            \makecell{Early\\Renaissance} &
            {Ukiyo-e} \\
            \includegraphics[width=.111\textwidth,height=.111\textwidth]{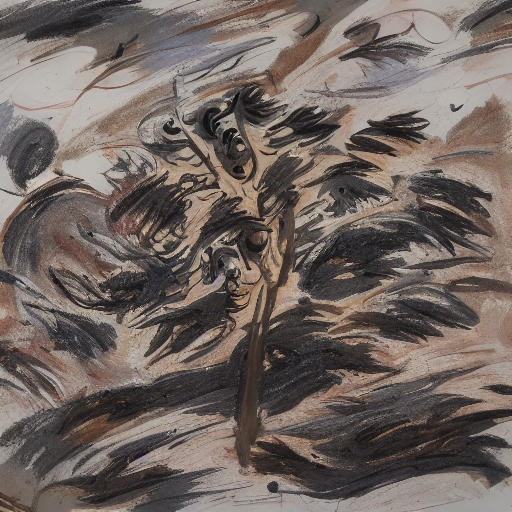} & % Mannerism (Late Renaissance)
            \includegraphics[width=.111\textwidth,height=.111\textwidth]{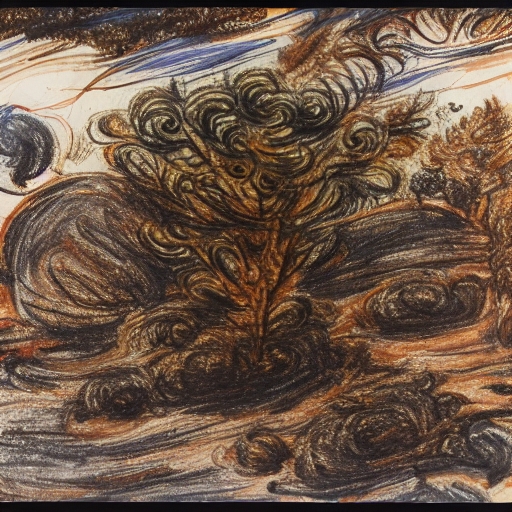} & % High Renaissance
            \includegraphics[width=.111\textwidth,height=.111\textwidth]{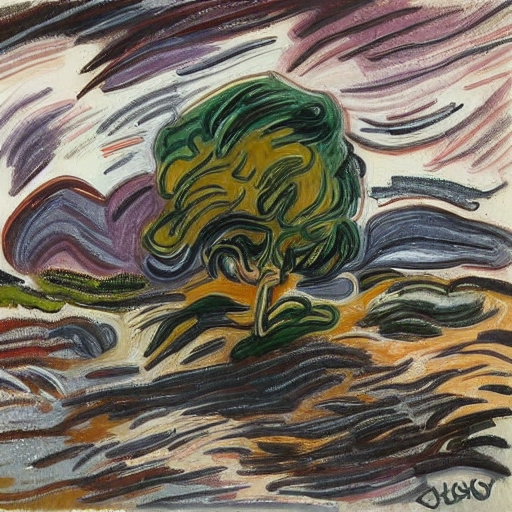} & % Fauvism
            \includegraphics[width=.111\textwidth,height=.111\textwidth]{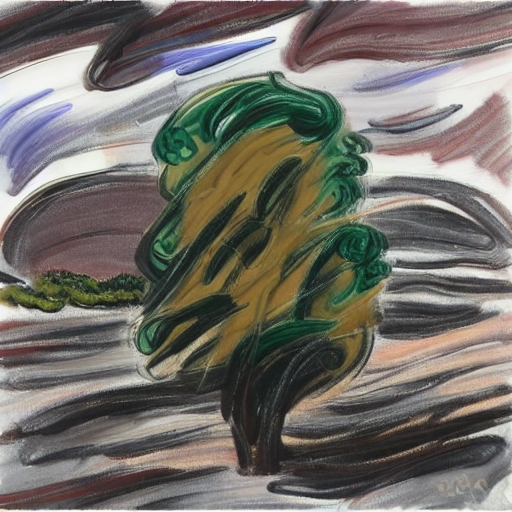} & % Minimalism
            \includegraphics[width=.111\textwidth,height=.111\textwidth]{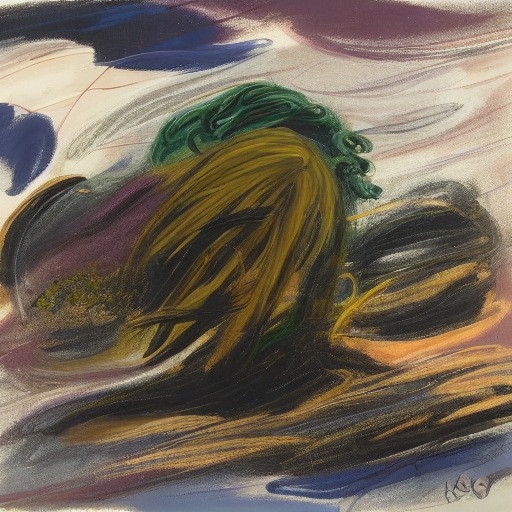} & % Action painting
            \includegraphics[width=.111\textwidth,height=.111\textwidth]{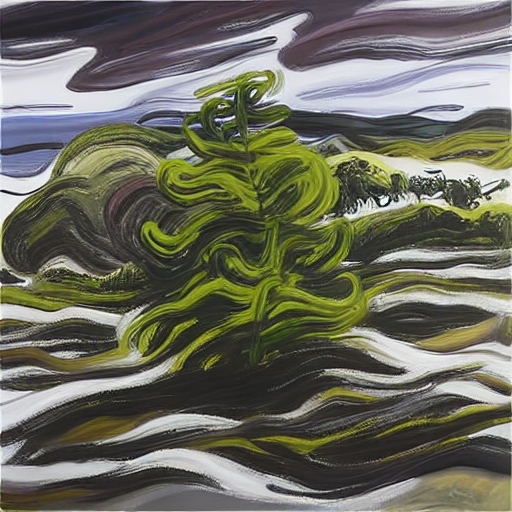} & % Contemporary Realism
            \includegraphics[width=.111\textwidth,height=.111\textwidth]{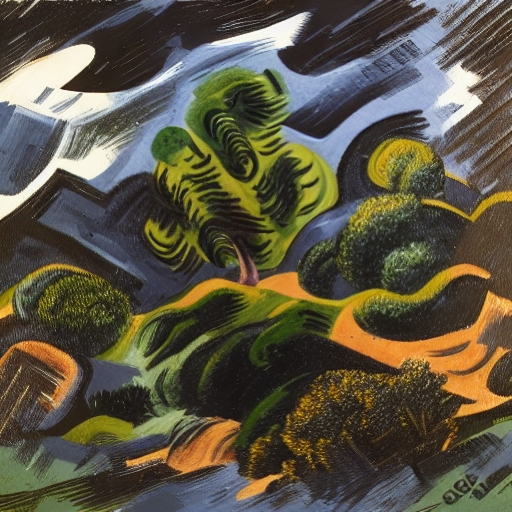} & % Synthetic Cubism
            \includegraphics[width=.111\textwidth,height=.111\textwidth]{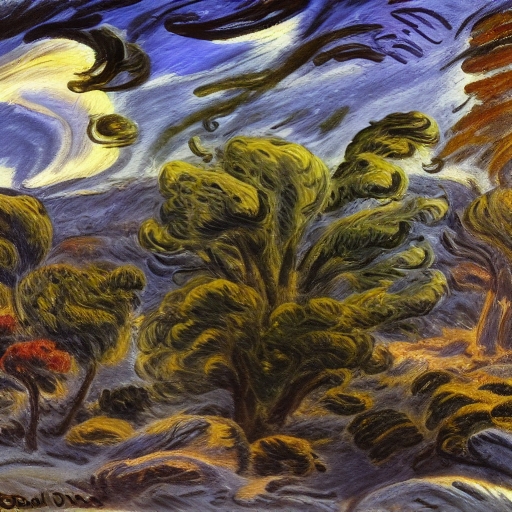} & % New Realism
            \includegraphics[width=.111\textwidth,height=.111\textwidth]{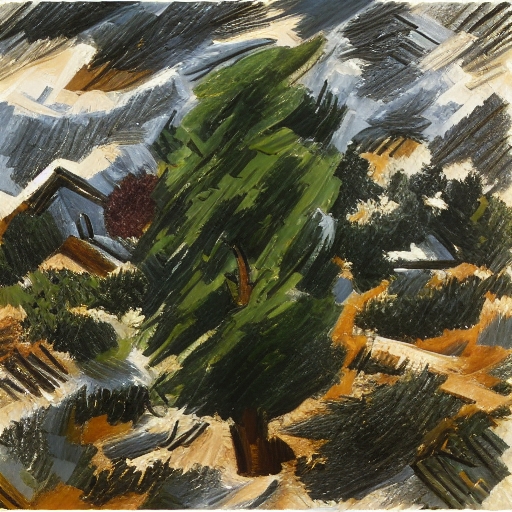}\\ % Analytical Cubism
            \makecell{Mannerism\\\teenyweeny(Late Renaissance)} &
            \makecell{High\\Renaissance} &
            {Fauvism} &
            {Minimalism} &
            {Action painting} &
            \makecell{Contemporary\\Realism} &
            \makecell{Synthetic\\Cubism} &
            {New Realism} &
            \makecell{Analytical\\Cubism} \\
        \end{tabular}
        \begin{minipage}[t]{\linewidth}
            \tiny
            \textbf{Prompt}: A landscape with a central tree surrounded by bushes and a background of swirling sky and distant trees.\\
            \textbf{PoA control}: \textbf{Asymmetric balance} is evident in the composition, with the central tree and surrounding bushes providing visual weight on one side, balanced by the distant trees and sky on the other side. This creates a dynamic yet stable composition. \textbf{Harmony} is achieved through the consistent use of swirling brushstrokes and a cohesive color palette, which unifies the various elements of the landscape.       \textbf{Variety} is present in the different types of vegetation and the varied brushstrokes, which add visual interest and complexity to the composition. \textbf{Unity} is achieved by the consistent style of brushwork and the natural theme, which ties all elements together into a coherent whole. \textbf{Contrast} is evident in the use of dark and light colors, particularly in the foliage and sky, which helps to highlight different areas and create depth. \textbf{Emphasis} is placed on the central tree, which stands out due to its size, position, and the detailed brushwork that draws the viewer's attention. \textbf{Proportion} is maintained with the central tree being the largest element, indicating its importance, while the surrounding bushes and distant trees are smaller, creating a sense of depth. \textbf{Movement} is created by the swirling brushstrokes in the sky and foliage, which guide the viewer's eyes across the composition and suggest a dynamic, flowing scene. \textbf{Rhythm} is established through the repetitive and flowing brushstrokes, which create a visual tempo and connect different parts of the composition. \textbf{Pattern} is present in the repetitive brushstrokes used for the foliage, which add texture and visual interest to the composition.
        \end{minipage}
    \end{subfigure}
    \vspace{-8pt}
    \caption{ArtDapted generations by fixing on the above conditions and varying across all 27 art-styles.}
    \squeezegap
    \label{fig:style-wise}
\end{figure}

We evaluate at \textit{two levels}: (i) \textit{Principle-level}, which identifies the winning model for each PoA condition using GPT scores, with ties broken by IR; and (ii) \textit{Composition-level}, which determines the overall winner based on the majority of principle-level wins, with ties broken by composition-level IR. The former provides granular measurement on alignment to individual PoA controls, while the latter provides overall measurement on adherence to multiple composition controls. For principle-level IR, the scoring prompt concatenates the contextual caption with the corresponding PoA control. For composition-level IR, the scoring prompt concatenates the caption with all PoA controls in fixed order.

Crucially, on every \textit{set} of art-conditions in the test set, we obtain the outputs from each participating model and tabulate their scores across both evaluation levels into a \textit{scorecard}. GPT is instructed to assign scores on a 7-point Likert scale (1 = poor, 7 = excellent). An example scorecard is shown in~\cref{fig:scorecard-example}.

We assess the consolidated results from each evaluation level under two \textit{assessment settings}: ASSMT and ASSMT*. In ASSMT*, the original CompArt artworks are also included in the competition pool, posing a much more challenging setting that benchmarks generated outputs against the ground-truth images.

\subsection{Quantitative results}
\label{sec:quantitiative_results}

\begin{figure}[t]
  {\tiny
    \def\colw{.19\linewidth}
    \def\imgw{1\linewidth}
    \def\txtw{1\linewidth}
    \setlength{\columnsep}{0pt}
    %
    % --- ROW 1 ---
    \begin{minipage}[t]{\colw}
        \centering
        \includegraphics[width=\imgw]{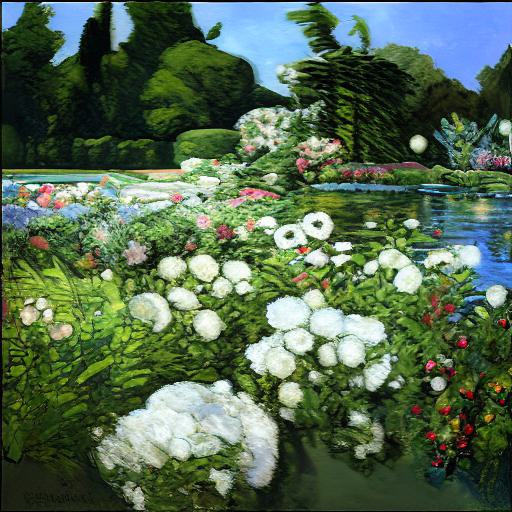}
        \parbox[t]{\txtw}{\raggedright
        \textbf{Asymmetric balance} is evident in the composition, with the dense cluster of flowers on the right side balanced by the open space and water on the left, creating a dynamic yet stable visual experience.}
    \end{minipage}\hfill
    \begin{minipage}[t]{\colw}
        \centering
        \includegraphics[width=\imgw]{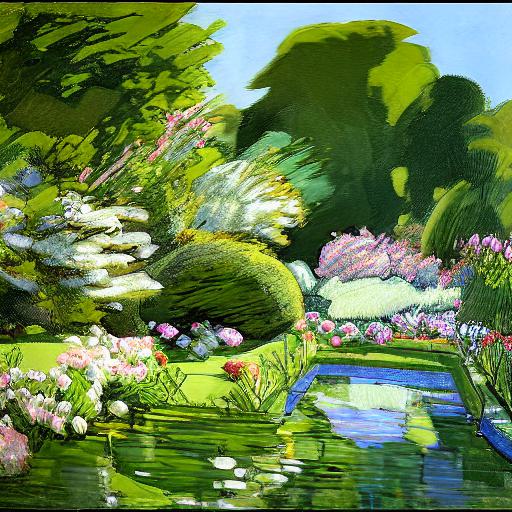}
        \parbox[t]{\txtw}{\raggedright
        \textbf{Harmony} is achieved through the consistent use of green and white hues, creating a cohesive and unified appearance that enhances the tranquil and serene atmosphere of the composition.}
    \end{minipage}\hfill
    \begin{minipage}[t]{\colw}
        \centering
        \includegraphics[width=\imgw]{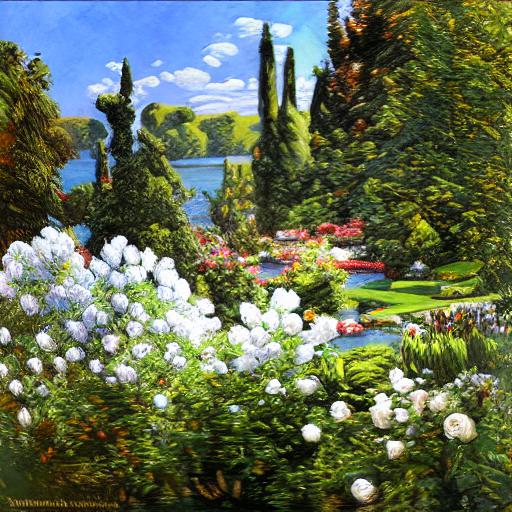}
        \parbox[t]{\txtw}{\raggedright
        \textbf{Variety} is present in the different types of foliage and flowers, as well as the varying shades of green, which add visual interest and prevent the composition from becoming monotonous.}
    \end{minipage}\hfill
    \begin{minipage}[t]{\colw}
        \centering
        \includegraphics[width=\imgw]{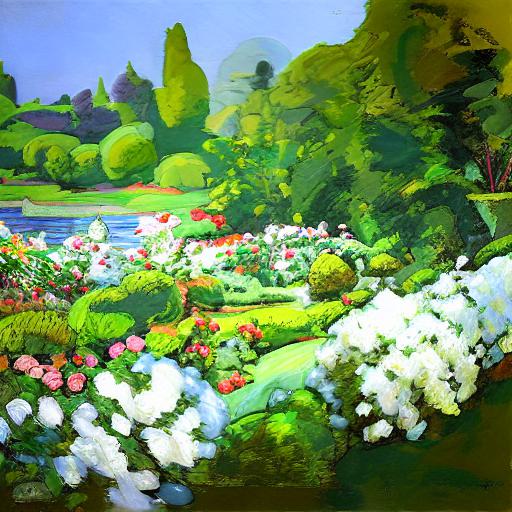}
        \parbox[t]{\txtw}{\raggedright
        \textbf{Unity} is evident in the composition through the consistent theme of a garden scene, with all elements contributing to the overall depiction of a lush and vibrant natural setting.}
    \end{minipage}\hfill
    \begin{minipage}[t]{\colw}
        \centering
        \includegraphics[width=\imgw]{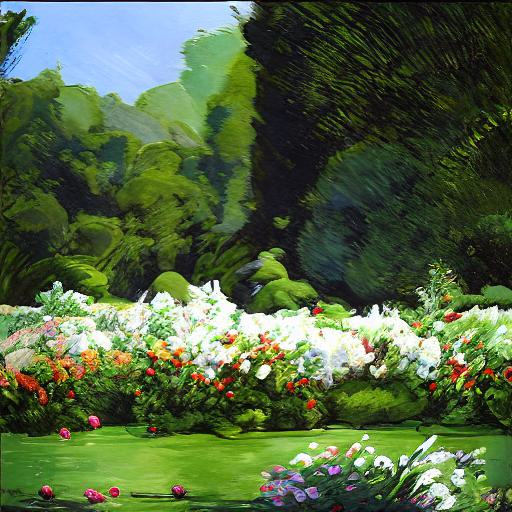}
        \parbox[t]{\txtw}{\raggedright
        \textbf{Contrast} is present in the composition through the juxtaposition of the dark green foliage against the bright white flowers, which helps to highlight the flowers and draw attention to them.}
    \end{minipage}
    \vspace{1em} %
    %
    % --- ROW 2 ---
    \begin{minipage}[t]{\colw}
        \centering
        \includegraphics[width=\imgw]{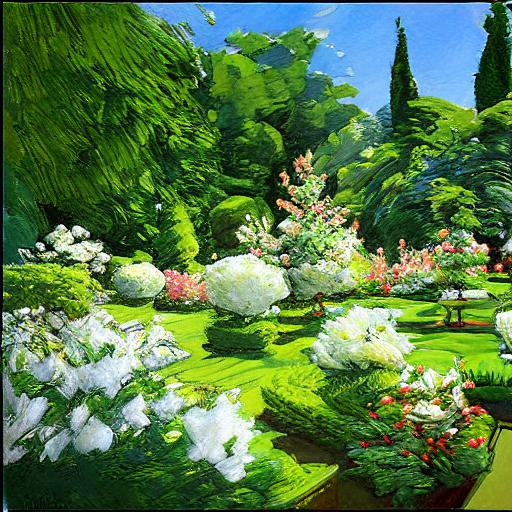}
        \parbox[t]{\txtw}{\raggedright
        \textbf{Emphasis} is placed on the blooming white flowers, which stand out against the green background and serve as focal points within the composition.}
    \end{minipage}\hfill
    \begin{minipage}[t]{\colw}
        \centering
        \includegraphics[width=\imgw]{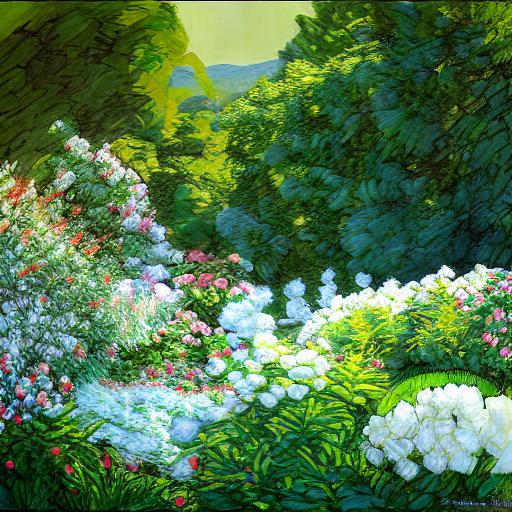}
        \parbox[t]{\txtw}{\raggedright
        \textbf{Proportion} is maintained with the relative sizes of the flowers and foliage, creating a realistic and believable garden scene that enhances the overall harmony and unity of the composition.}
    \end{minipage}\hfill
    \begin{minipage}[t]{\colw}
        \centering
        \includegraphics[width=\imgw]{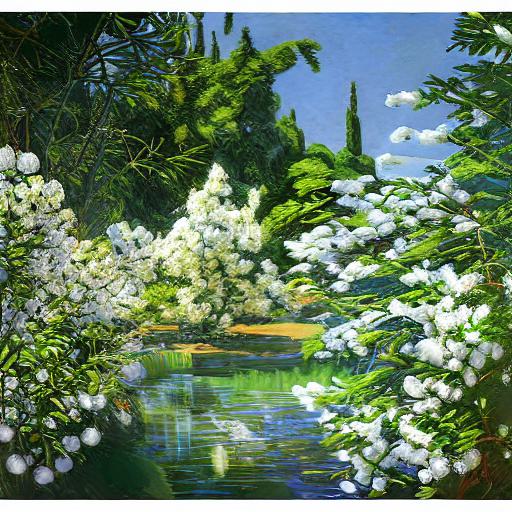}
        \parbox[t]{\txtw}{\raggedright
        \textbf{Movement} is suggested by the arrangement of the foliage and the direction of the branches, guiding the viewer's eye through the composition and creating a sense of flow and dynamism.}
    \end{minipage}\hfill
    \begin{minipage}[t]{\colw}
        \centering
        \includegraphics[width=\imgw]{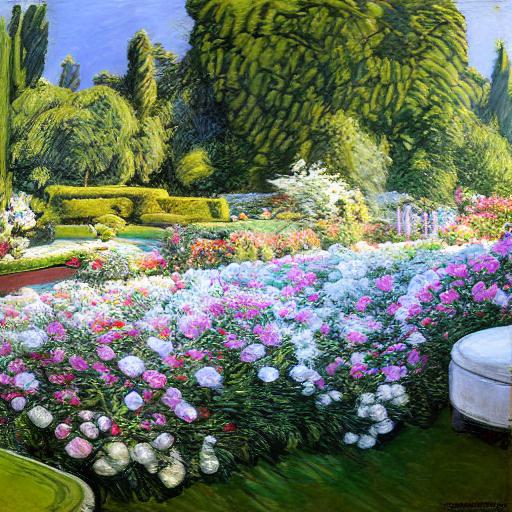}
        \parbox[t]{\txtw}{\raggedright
        \textbf{Rhythm} is created by the repetition of the white flowers and green foliage, which establishes a visual tempo and contributes to the overall harmony and unity of the composition.}
    \end{minipage}\hfill
    \begin{minipage}[t]{\colw}
        \centering
        \includegraphics[width=\imgw]{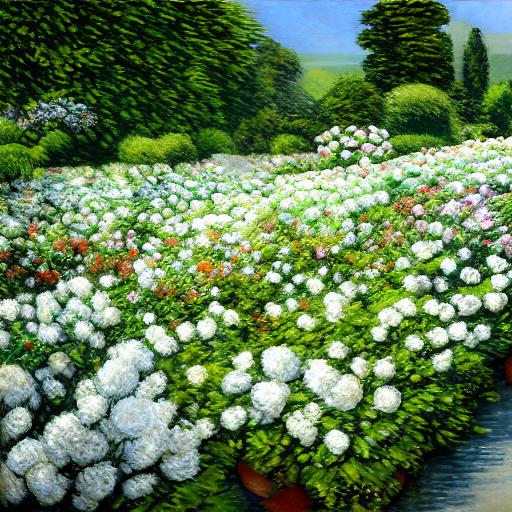}
        \parbox[t]{\txtw}{\raggedright
        \textbf{Pattern} is present in the consistent arrangement of the flowers and leaves, adding a sense of order and structure to the composition while enhancing its visual appeal.}
    \end{minipage}
    % --- BOTTOM TABLE (Prompt & Style) ---
    \begin{minipage}[t]{\linewidth}
         \textbf{Prompt}: A lush garden with abundant white flowers and green foliage, with a body of water and trees in the background.\\
         \textbf{Art-style}: Impressionism\\
    \end{minipage}
  }
  \squeezegap
  \caption{ArtDapted generations by fixing on the prompt and art-style and varying across each individual principle.}
  \squeezegap
  \label{fig:principle-wise}
\end{figure}

\cref{fig:evaluation-results} summarizes the results. In the standard assessment (ASSMT), ArtDapter consistently outperforms baselines in both principle-wise and composition-wise alignment by large margins, proving its effectiveness in bridging between high-level PoA concepts and diffusion guidance.

Notably, ArtDapter's superiority over ELLA—despite sharing the same TSC architecture—highlights that \textit{Aesthetic Alignment} is distinct from \textit{Semantic Alignment}. While ELLA excels at object placement, it lacks the specific vocabulary to handle compositional instructions. In the challenging ASSMT* setting (benchmarking against real artworks), although ArtDapter did not surpass the original CompArt artworks at the composition-level, it outperformed them at the principle-level across half of the PoA principles, demonstrating strong controllability. It is also noteworthy that in both levels of evaluation, ELLA consistently scored the worst.

Because FID relies on ImageNet-trained Inception features, its alignment with perceptual quality is limited in the art domain. Nevertheless, we investigate FID scores (evaluated at 256x256 resolution) to provide an additional quantitative reference from the perspective of perceptual statistics. We first split CompArt into two halves, $\mathrm{split}_1$ and $\mathrm{split}_2$, of sizes 40,017 and 40,018 respectively. We report the intra-dataset $\mathrm{FID}(\mathrm{split}_1,~\mathrm{split}_2)=1.54$. As our test set only comprises 1000 samples, it would exhibit much higher estimated variance than the CompArt splits. To address this, we expand the test set by generating outputs under single-PoA controls (i.e., principle-wise generations), yielding 8,979 outputs per model. We report the following scores: $\mathrm{FID}(\mathrm{ArtDapter},~\mathrm{split}_2)=27.13$, $\mathrm{FID}(\mathrm{ELLA},~\mathrm{split}_2)=21.00$, $\mathrm{FID}(\mathrm{SDv1.5},~\mathrm{split}_2)=29.27$. Interestingly, this suggests that ELLA is the most effective at producing ``artwork-like'' outputs among the three models. Importantly, this confirms that our evaluation scheme does not bias towards artistic appearances, but instead goes beyond perceptual similarity to assess compositional alignment.

\subsection{Qualitative analysis}
\label{sec:qualitative_analysis}

A core contribution of ArtDapter is the ability to \textit{disentangle} aesthetic composition (PoA) from Art-Style and Semantic Content. We observe this capability through two key behaviors:

\textbf{Style vs. Composition Independence.} As seen in \cref{fig:style-wise}, fixing the PoA controls while varying the Art-Style results in images that maintain the same structural composition (e.g., object placement, balance) but drastically change in texture and brushwork. For example, changing the style to ``Post-Impressionism'' while maintaining ``Rhythm'' introduces the characteristic swirling brushstrokes of Van Gogh, yet the underlying rhythmic flow of the scene remains consistent with the PoA prompt. This affirms ArtDapter's ability in decoupling the "how" (style) from the "structure" (PoA).

\textbf{Compositional Control.} Conversely, varying PoA conditions while fixing the prompt and style results in structural changes without altering the subject matter. This could be seen in~\cref{fig:principle-wise}. For example, the Variety principle provides the most diverse flora and colors while for the Harmony principle, the flora is depicted with the most consistent color palette, all while preserving the semantic subject of a lush garden in impressionism style. While the output variations for the Rhythm and Pattern principles depicted mostly similar looking scenery and perspectives, the latter one exhibited much more regularity and structure in its arrangement and use of white flowers.

\textbf{Threats to validity.}
Our primary metric uses the same model family that produced the pseudo-labels, which may favor outputs that match the annotation style. We mitigate this by (i) evaluating against strong baselines under identical prompts, (ii) including CompArt ground-truth artworks in ASSMT* as a harder reference point, and (iii) reporting ImageReward as an external signal. Nevertheless, absolute compositional quality remains subjective, and future work should incorporate expert or crowd studies when feasible.

\section{Related works}

\textbf{Text-to-Image Diffusion Models}~\cite{Ding2021CogViewMT,  Ramesh2021ZeroShotTG, Yu2022ScalingAM} are concerned with the conditional generation of an image output given a text prompt. Early approaches mostly depended on Generative Adversarial Networks (GANs)~\cite{DBLP:conf/nips/GoodfellowPMXWOCB14} which were prone to unstable training and exhibited weak generalization capabilities to open-domains~\cite{zhao2023unicontrolnet}. The introduction of DMs \cite{DBLP:conf/nips/HoJA20, DBLP:conf/iclr/SongME21} demonstrated that they did not suffer those limitations~\cite{dhariwal2021diffusion}. Seminal works such as Stable Diffusion~\cite{Rombach2021HighResolutionIS} and DALLE-2~\cite{DBLP:journals/corr/abs-2204-06125} ushered in a wave of developments~\cite{Brooks2022InstructPix2PixLT, DBLP:conf/iclr/FengHFJANBWW23, DBLP:conf/iclr/HertzMTAPC23} in the T2I task and laid the foundation for subsequent diffusion-based methods. The interest in additionally incorporating conditions from other modalities (e.g. depth/edge/segmentation maps, human poses, sketches etc.) motivated the design of popular lightweight adapters~\cite{DBLP:conf/aaai/MouWXW0QS24, DBLP:conf/iccv/ZhangRA23, DBLP:conf/cvpr/LiLWMYGLL23} that are attachable directly onto pre-trained DMs and trainable whilst keeping the DM's weights frozen. Subsequently, methods~\cite{DBLP:conf/icml/HuangC0SZZ23, zhao2023unicontrolnet, DBLP:conf/cvpr/WangX0CSWT22} were proposed to cater for multiple conditions in a composable manner and without having to train each condition independently.

\textbf{Semantic alignment} seeks to address the weakness of T2I DMs in adhering to complex prompts, especially pertaining to spatial relations and numeration constraints~\cite{Qu2023LayoutLLMT2IEL}. Common approaches to address this include adjusting cross-attention maps or embeddings to be more amenable towards the stated constraints~\cite{Chefer2023AttendandExciteAS, Li2023DivideB, Liu2022CompositionalVG} and fine-tuning T2I DMs using a reward mechanism built upon image-understanding feedback~\cite{fang2023boosting, Huang2023T2ICompBenchAC, Sun2025DreamSyncAT, DBLP:conf/nips/XuLWTLDTD23}. To overcome the limitations of CLIP~\cite{yuksekgonul2023when}, some instead leveraged the reasoning capabilities of Large Language Models (LLM)~\cite{DBLP:journals/corr/abs-2303-08774} to enhance the prompt or its embedding~\cite{10.5555/3666122.3669045, Zhong2023SURadapterET, Hu2024ELLAED} while others attempted to decompose prompts into multiple region constraints to provide fine and localised guidance for the T2I process~\cite{Qu2023LayoutLLMT2IEL, Cho2023VisualPF, DBLP:conf/cvpr/FengGCS0Z24, DBLP:journals/tmlr/LianLYD24, DBLP:conf/icml/0006YMXE024}.

\textbf{Art} has long been a prominent domain of study in the computer vision community, spanning a myriad of tasks ~\cite{Li2012RhythmicBD, Springstein2024VisualNL, Leslie2007OntologyBasedAO, 10.1167/jov.21.5.9, DBLP:conf/cvpr/KotovenkoWHO21, DBLP:conf/cvpr/ZouSQ0S21, DBLP:journals/corr/abs-1810-05977,DBLP:conf/icml/GaninKBEV18, DBLP:conf/iclr/HaE18}. Given the evocative nature of art, many works also focused on affective classification using image classification techniques~\cite{Machajdik2010AffectiveIC, Zhao2014ExploringPF}. Notably, \cite{Zhao2014ExploringPF} was the earliest to introduce PoA into the field and quantified them via hand-designed metrics. More recently, \cite{Achlioptas2021ArtEmisAL, Mohamed2022ItIO} demonstrated how neural speakers can be trained to reason about the emotions evoked by artworks.

\textbf{Artistic stylization} is closely-related to aesthetics because style can be seen as one aspect of aesthetics. The introduction of Image-to-image Translation and Neural Style Transfer~\cite{DBLP:conf/cvpr/GatysEB16} led to a wave of developments~\cite{DBLP:conf/iccv/ZhuPIE17, DBLP:conf/nips/ChenZWZZLXL21} and remains an active area of study~\cite{DBLP:conf/cvpr/DengTDMPWX22, DBLP:conf/cvpr/ZhangLLJZ22, DBLP:conf/eccv/FuWW22, DBLP:conf/cvpr/XuLN23, wu2023not, DBLP:conf/aaai/ZhangZXLZSLLHL24}. Stylization in the T2I domain falls under the broader task of Personalized T2I Synthesis. In exemplar-guided methods, reference images are used to optimize the text-encoder directly~\cite{DBLP:journals/corr/abs-2209-12330} or support text-inversion~\cite{DBLP:conf/iclr/GalAAPBCC23} to learn a new token embedding that can be used for stylistic or conceptual specification in the prompt. In parameter-efficient fine-tuning methods, a style-tuned model is trained for each style~\cite{sohn2023styledrop, DBLP:conf/iclr/HuSWALWWC22, DBLP:conf/nips/GuWWSCFXZCWGSS23}.
% Conclusion & future work: .25 page
% future work: indicate limitations
\section{Conclusion}

We introduced the problem of \textit{Aesthetic Alignment} in T2I generation, framing aesthetics not as an abstract quality but as a controllable set of compositional rules defined by the Principles of Art (PoA). To enable this line of research, we created \textbf{CompArt}, a large-scale dataset annotated with PoA, and proposed \textbf{ArtDapter}, a disentangled adapter that empowers diffusion models with visual literacy. Our evaluation demonstrates that formalizing aesthetics into a structured vocabulary allows for effective, user-driven control over image composition. Our code and data are publicly available.

\textbf{Limitations.} CompArt relies on model-generated analyses that may contain mistakes, especially for fine spatial relations or when resizing alters proportions. The dataset inherits biases from WikiArt's artist and style coverage, and PoA itself does not exhaust every aspect of visual composition. Finally, our main evaluation measures adherence to an operationalized rubric (plus an external preference model) rather than expert consensus; broader human studies remain an important direction for future work.

Although we target art generation here, the PoA-based approach is not limited to that domain: the same decomposition could be applied to photography and design, wherever compositional control is desirable. We hope it prompts further work toward bridging human creative intent and generative model capabilities.
\newpage
\bibliographystyle{ACM-Reference-Format}
\bibliography{references}
% \input{sec/auxiliary_materials}
\end{document}